\documentclass[12pt]{article}
\usepackage[utf8]{inputenc}
\usepackage[margin=2cm]{geometry}
\usepackage{fullpage,enumitem,amssymb,amsmath,amsthm,xcolor,cancel,gensymb,hyperref,graphicx,bbm}
\usepackage{caption}
\usepackage{subfigure}
\usepackage{comment}

\usepackage{indentfirst}

\usepackage{natbib}
\usepackage{algorithm}
\usepackage{algpseudocode}

\usepackage{hyperref}
\usepackage{cleveref}
\usepackage{setspace}
\usepackage{titlesec}
\titlespacing\section{0pt}{6pt plus 4pt minus 2pt}{0pt plus 2pt minus 2pt}
\titlespacing\subsection{0pt}{6pt plus 4pt minus 2pt}{0pt plus 2pt minus 2pt}
\titlespacing\subsubsection{0pt}{6pt plus 4pt minus 2pt}{0pt plus 2pt minus 2pt}

\newtheorem{thm}{Theorem}
\newtheorem{lem}{Lemma}
\newtheorem{rmk}{Remark}
\newtheorem{cor}{Corollary}
\newtheorem{prop}{Proposition}

\newtheorem{defn}{Definition}
\newtheorem{assm}{Assumption}
\crefname{lem}{Lemma}{Lemmas}

\DeclareMathOperator*{\argmax}{arg\,max}

\DeclareMathOperator*{\Var}{\text{Var}}

\DeclareMathOperator*{\bbE}{\mathbb{E}}
\DeclareMathOperator*{\bbR}{\mathbb{R}}

\DeclareMathOperator*{\bbP}{\mathbb{P}}
\DeclareMathOperator*{\calA}{\mathcal{A}}
\DeclareMathOperator*{\calB}{\mathcal{B}}

\DeclareMathOperator*{\calL}{\mathcal{L}}

\DeclareMathOperator*{\calS}{\mathcal{S}}
\DeclareMathOperator*{\calX}{\mathcal{X}}

\DeclareMathOperator*{\tr}{\textup{tr}}

\DeclareMathOperator*{\vect}{\textup{vec}}

\DeclareMathOperator*{\bbone}{{\mathbbm{1}}}

\newcommand{\wei}[1]{{\leavevmode\color{red}{[Wei: #1]}}}

\providecommand{\keywords}[1]
{
  \small	
  \textbf{\textit{Keywords---}} #1
}

\title{Low-Rank Contextual Reinforcement Learning from Heterogeneous Human Feedback}

\author{
    Seong Jin Lee\thanks{Ph.D. student, Department of Statistics and Operations Research, University of North Carolina, Chapel Hill. Email: slee7@unc.edu} \and
    Will Wei Sun\thanks{Associate Professor, Daniels School of Business, Purdue University. Email: sun244@purdue.edu} \and
    Yufeng Liu\thanks{Professor, Department of Statistics and Operations Research, Department of Genetics, Department of Biostatistics, The University of North Carolina at Chapel Hill. Email: yfliu@email.unc.edu.}
    \and
}

\date{}

\begin{document}

\maketitle

\onehalfspacing

\begin{abstract}

Reinforcement learning from human feedback (RLHF) has become a cornerstone for aligning large language models with human preferences. However, the heterogeneity of human feedback, driven by diverse individual contexts and preferences, poses significant challenges for reward learning. To address this, we propose a Low-rank Contextual RLHF (LoCo-RLHF) framework that integrates contextual information to better model heterogeneous feedback while maintaining computational efficiency. Our approach builds on a contextual preference model, leveraging the intrinsic low-rank structure of the interaction between user contexts and query-answer pairs to mitigate the high dimensionality of feature representations. Furthermore, we address the challenge of distributional shifts in feedback through our Pessimism in Reduced Subspace (PRS) policy, inspired by pessimistic offline reinforcement learning techniques. We theoretically demonstrate that our policy achieves a tighter sub-optimality gap compared to existing methods. Extensive experiments, ranging from synthetic simulations to an analysis of the real-world PersonalLLM benchmark, validate the effectiveness of LoCo-RLHF and demonstrate its superior performance in personalized settings.

\end{abstract}
\keywords{Distribution Shift, Large Language Models, Low-rankness, Offline Reinforcement Learning}

\newpage
\doublespacing
\setlength{\abovedisplayskip}{6pt minus 2pt}
\setlength{\belowdisplayskip}{6pt minus 2pt}
\setlength{\abovedisplayshortskip}{6pt minus 2pt}
\setlength{\belowdisplayshortskip}{6pt minus 2pt}

\baselineskip=23.5pt

\section{Introduction}

Reinforcement Learning (RL) is a popular framework where an agent learns to optimize a policy that maximizes the expected rewards, reflecting the success of completing a task. In standard RL, the reward function plays a critical role in guiding the agent's behavior. When a reward function is well-defined, RL can effectively solve problems \citep{mnih2015human, pmlr-v48-mniha16, silver2016mastering}. However, in many real-world applications, designing or observing an appropriate reward function can be a significant challenge. For instance, consider the problem of designing a reward model for a self-driving car. One might construct a heuristic reward structure, assigning a $+1$ reward for every mile traveled to encourage forward movement and a $-10$ penalty for traffic violations to promote safety. While this approach might seem straightforward, it can lead to unintended consequences. The car may prioritize speed over adherence to traffic regulations, misaligning with human safety preferences. Conversely, excessive penalties for traffic violations could result in the car avoiding movement altogether, merely to minimize penalties, disregarding its primary objective of safe driving. Such scenarios illustrate how traditional RL, dependent on fixed reward structures, often fails to capture the nuanced, human-aligned goals that are critical in complex environments such as autonomous driving \citep{sallab2017deep, kiran2021deep}.

Reinforcement Learning from Human Feedback (RLHF) overcomes this issue by incorporating human input into the learning process \citep{christiano2017deep, ziegler2019fine, bai2022training}. Unlike standard RL, where agents learn from predefined reward functions, RLHF uses human feedback to train a reward model that reflects human preferences for state-action pairs. RLHF is often referred to as preference-based reinforcement learning \citep{zhan2023provable} because it enables agents to learn from non-numerical feedback, which is more natural in many real-world settings. For example, when evaluating responses to a question, humans typically find it easier to compare answers and determine which is better, rather than assigning precise numerical scores. To capture these preferences, models like the Bradley-Terry-Luce (BTL) model \citep{bradley1952rank} are employed. The BTL model assumes that internal preferences influence action choices, with the probability of selecting one action over another being proportional to its preference. Explicitly, the probability of choosing action $a_1$ over $a_0$ is given by:
$$
\bbP(a_1 > a_0) = \frac{\exp(r(s, a_1))}{\exp(r(s,a_1)) + \exp(r(s,a_0))},
$$
where $r(s,a)$ represents the preference of action $a$ in state $s$. By learning these preferences, RLHF uses them as rewards to train policies that maximize the learned reward, aligning agent behavior more closely with human preferences. This is especially valuable in complex domains like LLMs, where traditional reward functions are difficult to design. After training a supervised model on large-scale text data, RLHF fine-tunes the model using human-derived reward signals to better align it with human preferences. This approach has been crucial in advancing large language models (LLMs) such as InstructGPT, ChatGPT and Llama \citep{ouyang2022training, achiam2023gpt, touvron2023llama}.

\begin{figure}[h!]
    \centering
    \captionsetup{width=.9\linewidth}
    \includegraphics[width = 0.7\textwidth]{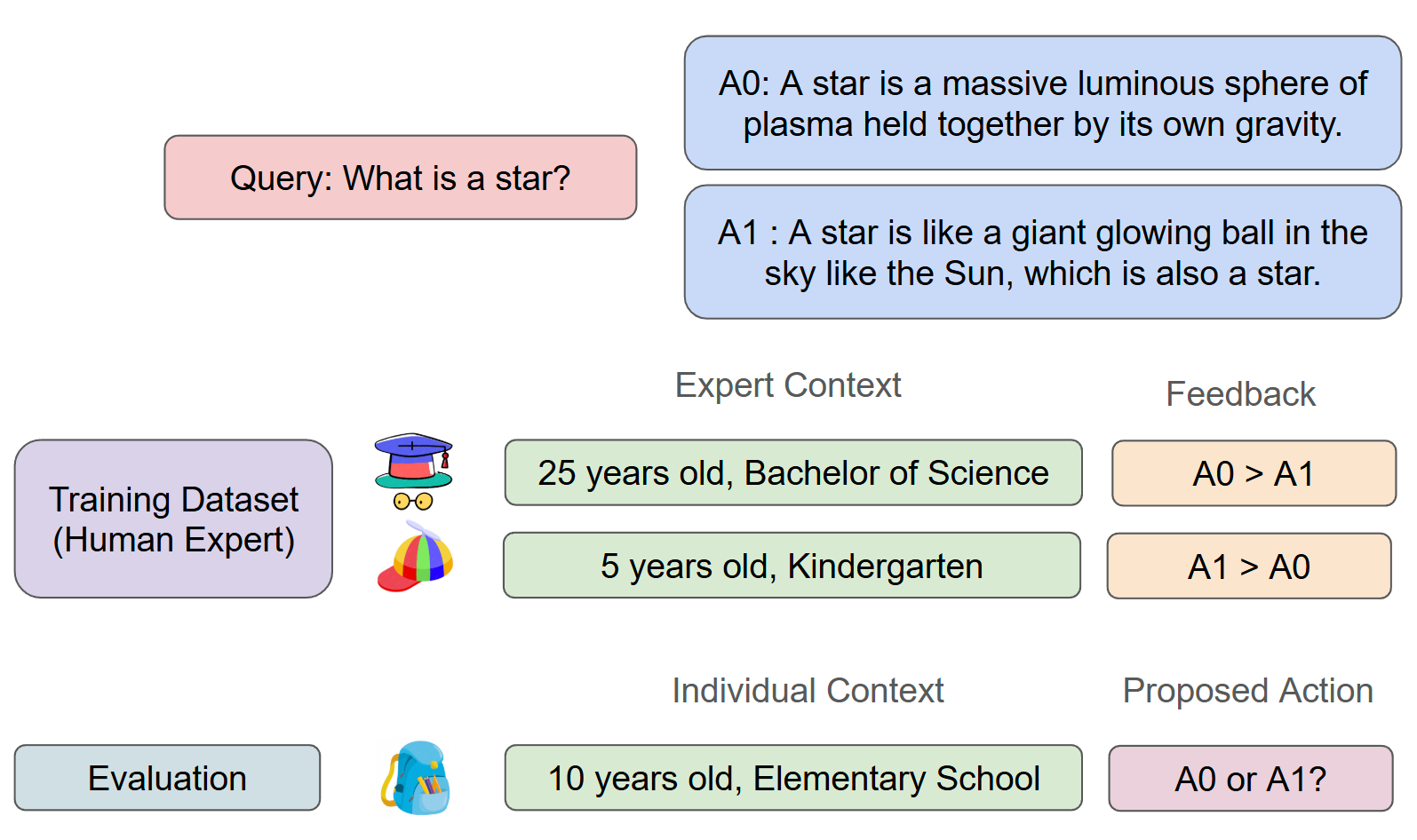}
    \caption{Illustration of personalization in RLHF.}
    \label{fig:fig_illust_personalization}
\end{figure}

Despite its successes, the existing RLHF framework faces notable challenges \citep{casper2023open}. As the reward model is a function of the query-answer pair, it assumes that all individuals share a single preference function. In practice, however, individuals exhibit diverse characteristics, leading to heterogenous preferences \citep{zhong2024provable,park2024rlhfheterogeneousfeedbackpersonalization}. 
The heterogeneity in human preferences can lead to three types of challenges. The first challenge is the \textbf{personalization} problem. For instance, consider a model tasked with answering the question, “What is a star?”. The model could offer a detailed scientific explanation, such as “A star is a massive, luminous sphere of plasma held together by its own gravity,” or a simpler response like “A star is a giant glowing ball in the sky, like the Sun.” While a scientifically inclined user might favor the former, a five-year-old child is more likely to prefer the latter for its simplicity and interpretability (See Figure \ref{fig:fig_illust_personalization}). This example illustrates the benefits of development of personalized models that adapt to user contexts, highlighting the potential of contextual RLHF to address these diverse needs.

\begin{figure}[h!]
    \centering
    \captionsetup{width=.9\linewidth}
    \includegraphics[width = 0.7\textwidth]{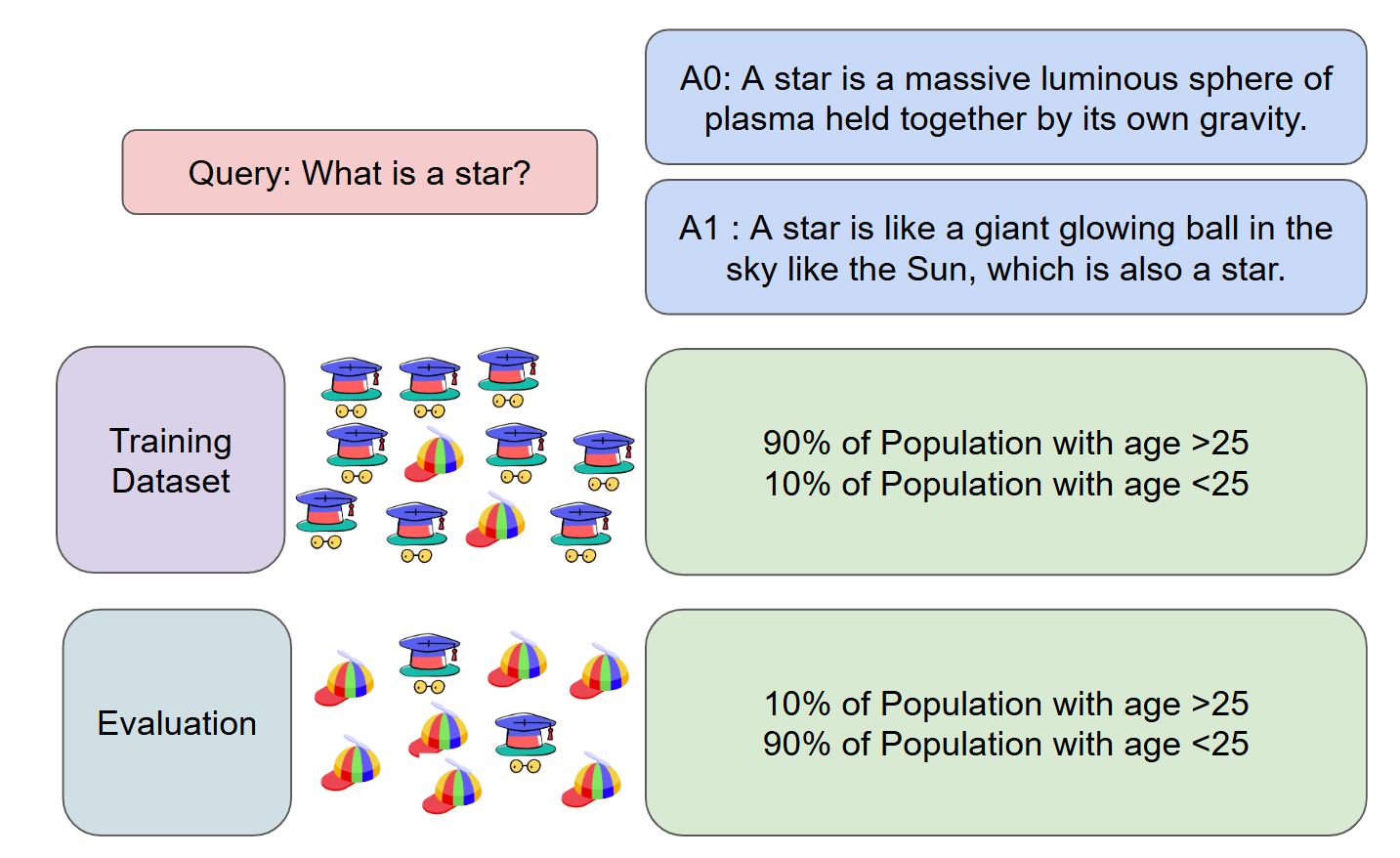}
    \caption{Illustration of distribution shift in RLHF.}
    \label{fig:fig_illust_distshift}
\end{figure}

The second challenge is the \textbf{distribution shift} between the offline training data and the target deployment distribution. For example, consider a scenario where offline feedback data are predominantly collected from college students, but the model is intended for deployment among preschool children as illustrated in Figure \ref{fig:fig_illust_distshift}. As the model is trained on the offline dataset via empirical risk minimization, the estimated homogeneous model will lead to optimize its performance on the college students, which could perform poorly with children. In such cases, accounting for user-specific features becomes essential to effectively bridge the gap between the training and target distributions, ensuring the model performs well even on different populations.

The last challenge is the \textbf{high dimensionality} of contexts in RLHF problems. In LLMs, reward models are typically trained using features extracted from the final layer of a pre-trained supervised model. For instance, GPT-3 (175B) \citep{brown2020language} has 12,288 units in its last layer, while InstructGPT, which employs a 6B reward model (RM), has 4,096 units in its last layer \citep{ouyang2022training}. Additionally, \citet{ouyang2022training} includes contextual information about labelers (human experts), such as demographic attributes (e.g., age, gender, ethnicity, nationality, education level) and survey-based details. Including both user contexts and state-action features in the model leads to a rapidly expanding parameter space, driven by the high-dimensional interaction terms. This growth poses substantial computational challenges for efficient parameter estimation.

\begin{figure}[h!]
    \centering
    \captionsetup{width=.9\linewidth}
    \includegraphics[width = 0.7\textwidth]{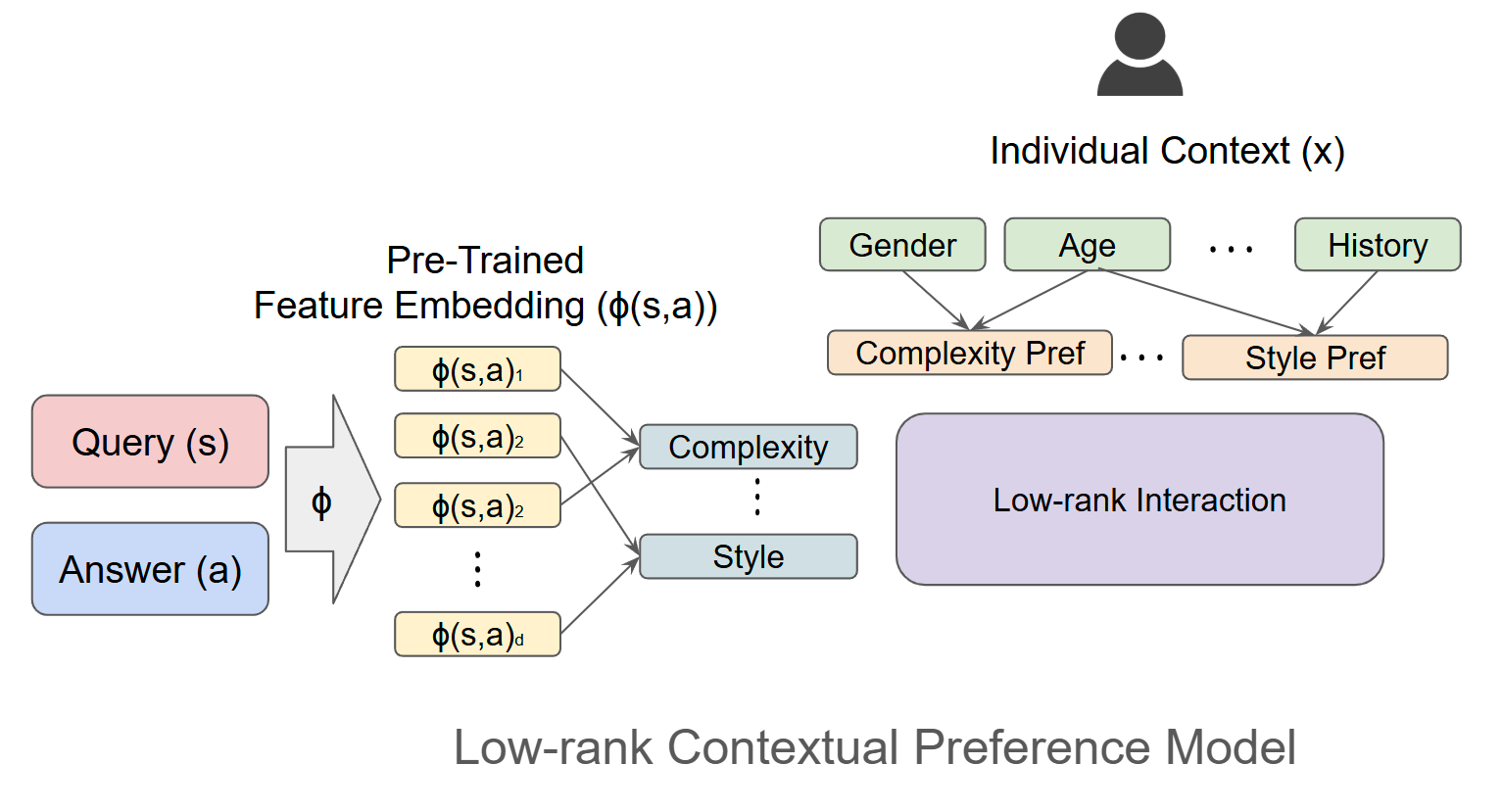}
    \caption{Illustration of the Low-rank Contextual Preference Model.}
    \label{fig:fig_illust_prefmodel}
\end{figure}

In this paper, we introduce a Low-rank Contextual RLHF (LoCo-RLHF) framework to address these challenges. Advances in digital platforms have enabled the collection of rich and diverse user data, which can be leveraged to develop individualized preference models. Instead of modeling user preferences as a homogeneous function of state-action pairs $r(s,a)$ for state $s$ and action $a$, we consider heterogeneous preference $r(x,s,a)$, where $x$ represents individual context. Specifically, the homogeneous feedback model considers $r(s,a) = \theta^\top \phi(s,a)$ for the pre-trained embedded feature $\phi(s,a) \in \bbR^{d_\phi}$ and some unknown vector parameter $\theta$. To allow heterogeneity induced from individual contexts, we adopt the bilinear form $r(x,s,a) = x^\top \Theta^* \phi(s,a)$, where $x\in \bbR^{d_x}$ and $\Theta^*\in \bbR^{d_x\times d_\phi}$ is a unknown matrix parameter. This formulation captures heterogeneity in preference by accounting for individual contexts. To address the high dimensionality, we leverage the low-rank structure of the parameter matrix. Low-rank approximations have been successfully applied in various domains, including contextual bandits \citep{jun2019bilinear, kang2022efficient} and LLMs \citep{li2018measuring, aghajanyan2020intrinsic, hu2021lora, ding2023parameter, dettmers2024qlora, sidahmed2024perl}. In this approach, high-dimensional features are projected into a low-dimensional latent space. For example, in the context of a query about a star, state-action features might include an embedding representing the complexity of the explanation, while user context features (e.g., age, education level) could capture preferences for simpler or more detailed responses. The interaction between these features and contexts is then modeled using a reduced set of latent factors. Motivated from this, we impose a low-rank structure on the matrix parameter $\Theta$ as shown in Figure \ref{fig:fig_illust_prefmodel} and hence substantially reduce the computation complexity from $d_x d_\phi$ to $(d_x + d_\phi) r$, where $r$ is the rank of the matrix. Details of the contextual preference model is given in Section \ref{subsect:ContextualPrefModel}. By imposing a low-rank structure, we effectively reduce the dimensionality of the parameter space, preserving the essential interactions while minimizing estimation errors and computational costs. 

We then design a Pessimism on Reduced Space (PRS) algorithm to solve this LoCo-RLHF problem. The PRS algorithm has the following three key components: estimating the low-rank subspace, constructing confidence bounds and pessimistic values, and deriving the pessimistic policy.

\begin{enumerate}
    \item \textbf{Estimating the low-rank subspace.}
    In the first stage, we leverage the low-rank structure of the parameter matrix to estimate a low-dimensional subspace using a rank-constrained maximum likelihood estimator. Since this optimization problem is inherently non-convex, we employ factored gradient descent (also known as alternating gradient descent) via the Burer-Monteiro formulation \citep{burer2003nonlinear, zheng2016convergence}. Once the matrix is estimated, we perform singular value decomposition (SVD) to extract orthogonal matrices that define the subspace. These matrices allow us to embed features and contexts into a reduced-dimensional space, thereby reducing the problem's overall complexity.
    \item \textbf{Constructing confidence bounds.}
    With the reduced subspace established, the next step involves projecting the original problem into the low-dimensional space and constructing confidence bounds within this representation. Subspace estimation inherently introduces uncertainty when projecting to the reduced rank. To address this, we develop a novel analytical framework that incorporates the uncertainty quantification into the confidence bound construction, ensuring robustness in subsequent stages.
    \item \textbf{Deriving the pessimistic policy.}
    Finally, we adopt a pessimistic approach, widely used in offline RL literature \citep{jin2021pessimism, fu2022offline, zhu2023principled, shi2024off}, to account for uncertainties in reward estimation. By optimizing the pessimistic reward, constructed to account for worst-case scenarios, we derive a robust policy that balances exploration and exploitation, even in the presence of distribution shift.
\end{enumerate}

In theory, we provide theoretical guarantees for our algorithm by deriving upper bounds on the sub-optimality gap. The sub-optimality gap is defined as the difference between the optimal reward, achieved with complete knowledge of the true parameters, and the reward of the policy derived by our algorithm. Specifically, we establish that the sub-optimality gap of our PRS policy is bounded as $O\Big(\sqrt{\frac{(d_x+d_\phi)\cdot r + \log(1/\delta)}{n}}\Big)$ with probability at least $1-\delta$ (Theorem \ref{thm:subopt}). In low-rank settings where $r \ll \min\{d_x, d_\phi\}$, our bound represents a significant improvement over the $O\Big(\sqrt{\frac{d_xd_\phi + \log(1/\delta)}{n}}\Big)$ bound achieved by existing methods \citep{zhu2023principled}. Furthermore, in the special case of our LoCo-RLHF framework where human feedback follows a pre-defined group structure, the derived bound aligns with the results of \citet{zhong2024provable}, demonstrating the tightness of our analysis.

The sub-optimality gap analysis of our policy presents several challenges. First the rank-constrained maximum likelihood estimation problem is inherently non-convex, which precludes the direct application of classical results from convex optimization literature. Moreover, the offline dataset comprises binary pairwise responses from human experts, rather than direct quantitative data. This characteristic complicates the derivation of the estimation bounds, as it requires handling the discrete nature of the data. During the uncertainty quantification step, both the estimation error of the subspace and the likelihood estimation contribute to the overall uncertainty of the parameter estimate in the reduced space. Lastly, the approximation of the preference function in the reduced space is similarly influenced by the subspace estimation error, which propagates through the analysis and affects the policy's value. To address these challenges, we develop novel tools that incorporate subspace estimation error into the construction of estimation bounds and the approximation of the preference function. 

Finally, we present extensive numerical experiments with varying rank, dimensionality, and offline data distributions to evaluate the proposed PRS policy, comparing it with the greedy policy and the pessimistic policy derived from the unconstrained MLE of \citet{zhu2023principled}. Furthermore, we conduct a systematic investigation into rank selection and perform sensitivity analyses on the confidence bound parameter and data split selection, with detailed results and discussions included in the Appendix. In addition to simulations, we apply our method to the PersonalLLM benchmark, a large-scale real-world dataset for preference-based RLHF with heterogeneous reward models. On both synthetic and real data, the PRS policy consistently achieves smaller sub-optimality gaps than existing methods, with particularly strong gains in low-rank and high-dimensional regimes. Moreover, experiments with artificially augmented noisy features demonstrate that PRS remains stable while baseline methods deteriorate, highlighting the robustness of our approach in realistic preference-learning settings.

\subsection{Related Work}

Our work is closely related to, yet distinct from, recent studies on RLHF with heterogeneous experts and low-rank LLMs.
\begin{itemize}
    \item \textbf{RLHF with Heterogeneous Experts:} The incorporation of human feedback into the RL framework, introduced by \citet{christiano2017deep}, has gained significant attention in recent years. Recent studies, such as \citet{ouyang2022training} and \citet{bai2022training}, highlight that experts may disagree on optimal responses, and biases in offline datasets can disadvantage underrepresented groups \citep{prabhakaran2021releasing, feffer2023moral, kirk2023personalisation}. To address the heterogeneity of human preferences, various approaches have been proposed. One common method is to impose a group structure on individuals and train separate reward models for each group, as seen in \citet{park2024rlhfheterogeneousfeedbackpersonalization}, and \citet{ramesh2024group}. These models are then aggregated using techniques such as social welfare functions, Pareto-optimal solutions \citep{boldi2024pareto}, or robust optimization \citep{chakraborty2024maxmin}. Another line of research focuses on personalized models tailored to individual preferences \citep{jang2023personalized, kirk2023personalisation, poddar2024personalizing}. However, most of these approaches lack theoretical guarantees for their performance. The most closely related work is \citet{zhong2024provable}, which proposes a multi-party RLHF framework that models and aggregates preferences across multiple individuals. While their method addresses heterogeneity through a pre-defined grouping structure, it requires a predetermined clustering of individual preferences and does not fully account for individual-specific contextual information. In contrast, our work extends this research by introducing a low-rank contextual preference model that incorporates expert characteristics and learns the latent low-rank representation of the reward model. 
    
    \item{\textbf{Low-rank LLMs:}} To improve LLM performance, many models leverage high-dimensional parameter spaces \citep{ouyang2022training, bai2022training}. As dimensionality increases, low-rank approximations have been proposed for fine-tuning \citep{hu2021lora, ding2023parameter, dettmers2024qlora}, supported by evidence of intrinsic low-dimensional structures within high-dimensional spaces \citep{aghajanyan2020intrinsic, sidahmed2024perl}. These models primarily focus on low-rank adoption during pre-training or fine-tuning. In contrast, our work addresses the heterogeneity of human feedback in the reward learning step of RLHF. Additionally, drawing from statistical theory on high-dimensional problems \citep{wainwright2019high}, we provide a rigorous theoretical analysis. While low-rank structures have been studied in bandit and RL problems \citep{jun2019bilinear, kang2022efficient, cai2023doubly, stojanovic2024model, duan2024online}, their application in RLHF remains underexplored.

    \item \textbf{Uncertainty Quantification and Statistical Inference:} Our reward modeling approach is also connected to the literature on uncertainty quantification for low-rank structures and ranking. Foundational works by \citet{chen2019inference} and \citet{chen2021bridging} established frameworks for inference and UQ in noisy matrix completion and robust PCA. Furthermore, our contextual BTL model aligns with recent studies on covariate-assisted ranking and MLE-based UQ, which also consider potential model misspecifications \citep{fan2024uncertainty, fan2024covariate}. While these existing works primarily focus on asymptotic confidence sets and statistical inference, our work provides a non-asymptotic analysis. This is essential for the offline RL setting, where finite-sample guarantees are required to establish the sub-optimality gap of the resulting policy.
\end{itemize}

In summary, our work presents the first provable low-rank contextual RLHF framework that tackles personalization, distribution shift, and high-dimensionality in heterogeneous human preferences.

\subsection{Notations}

We introduce the notations used throughout the paper. Let $[n]$ denote the set $\{1,2, \ldots, n\}$. The Frobenius norm of a matrix is denoted by $\|\cdot\|_F$, and the matrix-induced norm  $\|v\|_W = \sqrt{v^\top W v}$ is denoted by $\|\cdot\|_W$. The ball induced by the Frobenius norm, centered at $\Theta_0$ with radius $r$ is denoted by $\calB(\Theta_0, r) := \{\Theta: \|\Theta - \Theta_0\|_F \le r\}$. For a vector $v$, $(v)_i$ represents its $i$-th component. We write $\Sigma_1 \succ \Sigma_2$ if $\Sigma_1 - \Sigma_2$ is positive definite. Denote Kronecker product of two matrices as $\otimes$. 

\section{Problem Formulation}

In this section, we formally define the LoCo-RLHF problem within the framework of pairwise comparison feedback. Specifically, we (i) introduce the contextual preference model, which captures the heterogeneous preferences of individuals, and (ii) demonstrate how the contextual reward model addresses both the personalization and distribution shift challenges.

\subsection{Contextual Preference Model} \label{subsect:ContextualPrefModel}

To address the heterogeneity in human preferences regarding queries and their corresponding answers, we introduce the \textbf{contextual preference model}. Let $\calS$ denote the state (query) space, $\calA$ the action (answer) space, and $\calX$ the context space. For $t \in [N]$, a human expert with context $x_t \in \calX$ is presented with a state $s_t \in \calS$ and two actions, $a_{t0}, a_{t1} \in \calA$. The expert selects the preferred action, where $y_t = 1$ indicates a preference for $a_{t1}$ over $a_{t0}$, and $y_t = 0$ indicates a preference for $a_{t0}$. The offline dataset comprises a collection of (i) contexts, (ii) states, (iii) action pairs, and (iv) expert preferences, represented as ${(x_t, s_t, a_{t0}, a_{t1}, y_t) : t \in [N]}$ for each individual $t$.

The Bradley-Terry-Luce (BTL) model \citep{bradley1952rank} has been widely used to capture how preferences influence the choices of experts. Let $r_0$ and $r_1$ denote the preferences associated with actions $a_0$ and $a_1$, respectively. The BTL model assumes that the probability of selecting an action is proportional to the exponential of its preference. Specifically, the probabilities are given by:
$$
\bbP(y=1) = \frac{\exp(r_1)}{\exp(r_0) + \exp(r_1)} \quad \text{and} \quad \bbP(y=0) = \frac{\exp(r_0)}{\exp(r_0) + \exp(r_1)}.
$$

The homogeneous preference model \citep{ouyang2022training, achiam2023gpt, touvron2023llama} assumes that the probability of choosing an action depends solely on the query and the available actions, making it identical across individuals. To account for heterogeneity, we introduce a \textbf{contextual preference model}, which reflects individual-specific preferences. Specifically, we model the reward function using a bilinear form: 
\begin{equation} \label{eqn:reward} 
r_{\Theta^*}(x,s,a) = x^\top \Theta^* \phi(s,a), 
\end{equation} 
where $x \in \calX$ represents the individual's context, $\Theta^* \in \mathbb{R}^{d_x \times d_\phi}$ is a parameter matrix, and $\phi(s, a)$ is the feature embedding for the query-action pair. Under this model, the probability of selecting action $a_1$ over $a_0$, given query $s$ and context $x$, is:
$$
{\bbP}(y = 1 | x,s,a_1, a_0) = \frac{\exp(x^\top \Theta^* \phi(s,a_1))}{\exp(x^\top \Theta^* \phi(s,a_0)) + \exp(x^\top \Theta^* \phi(s,a_1))}.
$$
This formulation introduces variability in the choice probabilities by incorporating the context $x$, thereby capturing the heterogeneity among individuals. 

The high dimensionality of context and feature embeddings poses significant computational challenges, as estimating the matrix parameter scales with the product of their dimensions. Inspired by the successful application of low-rank approximations in reinforcement learning \citep{jun2019bilinear, hu2021lora}, we impose a low-rank structure on the parameter matrix. Let $r$ be the rank of the parameter matrix $\Theta^*$, and let $\Theta^* = U^* D^* (V^*)^\top$ represent its singular value decomposition (SVD) of $\Theta^*$, where $U^*$ and $V^*$ are square matrices, and $D^*\in \bbR^{d_x\times d_\phi}$ with $\{d_i:i \in [r]\}$ as its $r$ non-zero diagonal entries. The reward function $r_{\Theta^*}(x,s,a)$ can then be expressed as:
$$
r_{\Theta^*}(x,s,a) =x^\top \cdot U^* D^* (V^*)^\top \cdot \phi(s,a),
$$
which simplifies to:
$$
r_{\Theta^*}(x,s,a) = \sum_{i=1}^r d_i \left( (U^*)^\top x \right)_i \left( (V^*)^\top \phi(s,a) \right)_i.
$$
This formulation demonstrates that $U^*$ and $V^*$ project the original features into a low-dimensional subspace, significantly reducing computational complexity, as illustrated in Figure \ref{fig:fig_illust_prefmodel}.

\begin{rmk}[Model Assumptions in RLHF]
Our setting aligns with standard preference-based reward modeling in RLHF, where feedback is collected via pairwise comparisons and rewards are learned over high-dimensional representations produced by pre-trained models \citep{ouyang2022training, zhu2023principled, zhong2024provable}. Contextual information, when available, can be incorporated to account for systematic heterogeneity across users or tasks, as is common in large-scale human feedback datasets \citep{kirk2023personalisation}. The low-rank structure assumed in this work is motivated by empirical findings in RLHF and preference learning, which suggest that variation in human preferences is often governed by a small number of latent factors despite the high dimensionality of learned representations \citep{hu2021lora,park2024rlhfheterogeneousfeedbackpersonalization, bose2025lore, fan2025uncertainty}. 
\end{rmk}

\begin{rmk}[Flexibility of the bilinear model]
The bilinear reward model generalizes various existing reward models. First, when $d_x = 1$ with $\calX = \{1\} \in \bbR$, the model reduces to $r_\theta(s,a) = \theta^\top \phi(s,a)$, corresponding to the reward model for homogeneous individuals \citep{zhu2023principled}. Second, the Boltzmann-model \citep{jeon2020reward, barnett2023active} models the reward as $\beta r(s,a)$, where $\beta$ is a rationality parameter. If the reward $r(s,a)$ is linear with respect to $\phi(s,a)$, such that $r(s,a) = \theta_1^\top \phi(s,a)$, and $\beta$ is a linear function of context $x$ i.e., $\beta(x) = x^\top \theta_2$, then the reward function becomes:
$
r(x,s,a) = \beta(x) r(s,a) = x^\top \theta_2 \theta_1^\top \phi(s,a),
$
which is the special case of formulation (\ref{eqn:reward}) with $\Theta^* = \theta_2 \theta_1^\top$ being a rank-1 matrix. Finally, to model the diversity of human preferences, \citet{zhong2024provable} considers $M$ types of individuals each with reward given as $r_m = (\theta_m^*)^\top \phi(s,a)$, where $\theta_m^* \in \bbR^d$ and $\phi: \calS \times \calA \to \bbR^d$ is a known feature mapping. This setup is also a special case of formulation (\ref{eqn:reward}), where $M = d_x$, $d = d_\phi$, $\Theta^* = (\theta_1^*, \theta_2^*, \ldots, \theta_M^*)^\top$, and $x = e_m$ for individual of type $m$, where $e_m$ is the $m$-th standard unit vector in $\bbR^{M}$.
\end{rmk}

\subsection{Value of Policy}

Next, we show how the contextual preference model effectively tackles personalization and distribution shift challenges. Let $\rho$ be the joint distribution of context $x$ and state $s$ in the evaluation set. In the \textbf{personalization} problem, the individual feature $x$ is available for the policy, enabling the policy to adapt its responses based on the context of each individual. Formally, the policy is defined as $\pi: \calX \times \calS \mapsto \calA$, allowing it to provide tailored answers for different contexts. The value of the policy is then defined as:
$$
J(\pi) := {\bbE}_{(x,s) \sim \rho} r_{\Theta^*}(x,s, \pi(x,s)).
$$

In the personalization problem, the objective is to determine an individualized policy $\pi(x,s)$ that provides personalized responses for each individual with context $x$, even when the the same query $s$ is presented. The goal is to maximize the value of the policy. 

The second case is the \textbf{distribution shift} problem. Unlike the personalization problem, the features of individuals are unavailable during evaluation in this second case. 
In this case, the policy is defined as a function of $s$ alone, i.e., $\pi(s)$. The corresponding value is given by
$$
J(\pi) := {\bbE}_{(x,s) \sim \rho} r_{\Theta^*}(x,s, \pi(s)).
$$

In the distribution shift problem, the objective is to find a global policy $\pi(s)$ that provides a unified response for all individuals, maximizing the expected value of the policy. It is important to note that the evaluation distribution $\rho$ may differ from the offline training distribution. If the contextual preference model is not considered, a homogeneous reward model would optimize the policy to maximize value based on the offline data distribution, potentially leading to suboptimal performance in the evaluation environment.

For brevity, we use the notation $r_{\theta}(x,s,\pi)$ to represent $r_{\theta}(x,s,\pi(x,s))$ in the personalization problem and $r_{\theta}(x,s,\pi(s))$ in the distribution shift problem. The sub-optimality of a policy is defined as 
$$
\texttt{SubOpt}(\pi) := J(\pi^*) -J(\pi),
$$ 
where $\pi^* = \argmax_\pi J(\pi)$ denotes the optimal policy. The objective is to determine a policy that minimizes the sub-optimality by leveraging the offline dataset.

\begin{rmk}[Connection with \citet{zhong2024provable}]
When $\calX$ represents the set of standard unit vectors in $\bbR^M$ and $x\sim \rho_x$ with $\rho_x$ being the uniform distribution, the formulation aligns with the framework proposed by \citet{zhong2024provable} under the Utilitarian welfare objective. However, while their approach assumes that the query and answers $(s_i, a_{i0}, a_{i1})$ are generated from the same distribution for all individuals, our framework accommodates scenarios where different state-action pairs are presented to each human expert. Additionally, our framework can be extended to accommodate alternative objective functions. Specifically, the value function can be generalized as: $J(\pi) := {\bbE}_{(x,s) \sim \rho} l(r_{\Theta^*}(x,s,\pi(s))-r_0),$ where $l$ is a specified transformation, and $r_0$ represents the minimum value of the reward function. For example, when $l$ is the logarithmic function, this formulation corresponds to the Nash welfare objective. The theoretical results remain valid with slight modifications, provided that $l$ is a Lipschitz continuous function.
\end{rmk}

\section{Algorithm}

In this section, we propose the \textbf{Pessimism in Reduced Subspace (PRS)} policy to address the LoCo-RLHF problem. Before introducing the details, we first outline the three main steps in the PRS algorithm. 
\begin{enumerate}
    \item \textbf{Estimation of the Low-rank Subspace:} We begin by solving the rank-constrained maximum likelihood estimation problem to obtain the low-rank estimate $\hat\Theta$. Using the SVD $\hat\Theta = \hat{U} \hat{D} (\hat{V})^\top$, we obtain an estimate of the low-rank subspaces, where $\hat{U}$ and $\hat{V}$ represent the linear transformation into the low-rank subspace.

    \item \textbf{Reduction to the Low-rank Subspace:}
    Next, we apply the rotation-truncation-vectorization (RTV) process to the parameter $\Theta$, individual contexts and embedded features with respect to the estimated subspaces $\hat{U}$ and $\hat{V}$. This transformation reduces dimensionality of the parameter space from $d_x d_\phi$ to $(d_x+d_\phi)r - r^2$, while introducing minimal error from subspace estimation. Using the reduced space, we compute the MLE $\hat{\theta}_{rtv}$ of the approximate likelihood function, which is used to efficiently estimate the true preference function.
    
    \item \textbf{Pessimism in the Reduced Space:} Finally we construct the confidence set $\mathbf{\Theta}_B(\hat\theta_{rtv})$ around the estimate $\hat\theta_{rtv}$ by quantifying the uncertainty in the estimation process. Using this confidence set, we define the pessimistic value function:
    $$
    \hat{J}(\pi) := \min_{\theta \in \mathbf{\Theta}_B(\hat\theta_{rtv})} {\bbE}_{(x,s)\sim \rho} r_{\theta_{rtv}}(x,s,\pi),
    $$
    which represents the pessimistic estimate of the reward under a given policy. The pessimistic policy $\hat\pi$ is then defined by solving $\hat\pi = \argmax \hat{J}(\pi)$.
\end{enumerate}

\subsection{Estimation of the Low-rank Subspace}

We begin by splitting the dataset into two partitions: the first $N_0$ observations are used to estimate the low-rank subspace, while the remaining $N-N_0$ observations are used to estimate the parameter and quantify the uncertainty in the reduced space. Given the dataset $\{(x_t, s_t, a_{t0}, a_{t1}, y_t : t\in [N_0]\}$, the contextual preference model in the BTL model (\ref{eqn:reward}) implies that the negative log-likelihood is given by:
\begin{equation} \label{eqn:likelihood}
\calL(\Theta) = \frac{1}{N_0} \sum_{t=1}^{N_0} \left( \log \left( 1 + \exp(-x_t^\top \Theta (\phi(s_t, a_{t1}) - \phi(s_t, a_{t0}))) \right) + (1-y_t) x_t^\top \Theta (\phi(s_t, a_{t1}) - \phi(s_t, a_{t0})) \right).
\end{equation}

Based on it, the rank-constrained optimization problem can be formulated as
$$
\min_{\Theta \in \bbR^{d_x \times d_\phi}, \text{rank}(\Theta)\le r} \mathcal{L}(\Theta),
$$

which can be solved using the Burer-Monteiro formulation \citep{burer2003nonlinear}. This approach factorizes $\Theta = UV^\top$, where $U\in \bbR^{d_x\times r}$ and $V\in \bbR^{d_\phi\times r}$. To ensure identifiability, we introduce a regularization term following \citet{zheng2016convergence}. The resulting optimization problem is formulated as:
$$
\min_{U\in \bbR^{d_x\times r}, V \in \bbR^{d_\phi\times r}} \left( \tilde{\calL}(U,V) := \calL(UV^\top) + \frac{1}{8} \|U^\top U - V^\top V\|_F^2 \right).
$$

To solve this non-convex optimization problem, we employ the alternating Factored Gradient Descent (FGD) method. FGD performs gradient descent on each component iteratively:
$$
U \leftarrow U - \eta \cdot \nabla_U \tilde{\calL}(U,V), \quad V \leftarrow V - \eta \cdot \nabla_V \tilde{\calL}(U,V),
$$
where $\eta$ denotes the learning rate, and $\nabla_U, \nabla_V$ are the gradients with respect to $U$ and $V$, respectively. By alternately updating each component, the estimated parameter converges to the true parameter, provided that certain conditions on the initial estimate and the learning rate are satisfied \citep{zheng2016convergence, zhang2023generalized}. The full details of the FGD procedure are outlined in Algorithm \ref{alg:fgd}.

\begin{algorithm}
 \caption{FGD - Factored Gradient Descent}
 \label{alg:fgd}
 \begin{algorithmic}[1]
 \Require Gradients of the loss function $\nabla_U\calL$ and $\nabla_V\calL$, initial estimate $\Theta_0$ and step size $\eta$.
 \Ensure SVD $\Theta_0 = U_0 D_0 V_0^\top$ and initialize $U^{(0)} = U_0D_0^{1/2}$ and $V^{(0)} = V_0D_0^{1/2}$.
 \For{$k=1, \ldots$,}
 \State Update $U^{(k+1)} = U^{(k)} - \eta \nabla_U {\calL}|_{U = U^{(k)}, V = V^{(k)}}$;
 \State Update $V^{(k+1)} = V^{(k)} - \eta \nabla_V {\calL}|_{U = U^{(k+1)}, V = V^{(k)}}$;
 \State Repeat until objective function converges.
 \EndFor
 \end{algorithmic}
\Return $\hat\Theta = U^{(k)} (V^{(k)})^\top$.
\end{algorithm}

In our experiments, we compute the initial estimate $\Theta_0$ using the unconstrained maximum likelihood estimator. Convex optimization algorithms, such as L-BFGS \citep{liu1989limited}, or gradient-based methods can be applied to obtain this initial parameter estimate.

\subsection{Reduction to the Low-rank Subspace}

RL algorithms leverage uncertainty quantification to improve performance, such as the ``optimism in the face of uncertainty" approach for online RL \citep{abbasi2011improved} and pessimistic policies for offline RL \citep{jin2021pessimism, rashidinejad2021bridging}. Existing uncertainty quantification methods are primarily designed for vector parameters, making them less suited to our high-dimensional matrix parameter space $\bbR^{d_x \times d_\phi}$. The large number of components in this space as well as the low-rank structure of the matrix parameter create significant challenges for effective uncertainty quantification. To address this challenge, we propose the ``rotation-truncation-vectorization" (RTV) method. This mapping explicitly exploits the low-rank structure of $\Theta^*$ to reduce the effective dimension from $d_x d_\phi$ to $r(d_x+d_\phi)$. The construction proceeds in three steps: aligning the matrix via rotation, discarding negligible components via truncation and flattening the result for inference. 

Let $\hat\Theta = \hat{U} \hat{D} \hat{V}^\top$ be the SVD of the estimate obtained from Algorithm \ref{alg:fgd}. Intuitively, if the estimated subspaces $\hat{U}$ and $\hat{V}$ are accurate, rotating the true parameter $\Theta^*$ in the direction of $\hat{U}$ and $\hat{V}$ should yield a close approximation of the rank $r$ diagonal matrix. 
For a detailed explanation, let $(U^*_1, U_2^*)$, $(V_1^*, V_2^*)$, $(\hat{U}_1, \hat{U}_2)$ and $(\hat{V}_1, \hat{V}_2)$ be partitions of $U^*, V^*, \hat{U}$ and $\hat{V}$, where $(\cdot)_1$ represent the first $r$ columns and $(\cdot)_2$ denotes the remaining columns of the matrix. Let $\Theta^*_r$ be the rotation of $\Theta^*$ with respect to the estimated subspace given by $\Theta^*_r := \hat{U}^\top \Theta^* \hat{V}$. We partition $\Theta_r^*$ into four sub-blocks based on the rank $r$:
$$
\Theta_r^* = 
\begin{pmatrix} 
(\Theta_r^*)_{11} & (\Theta_r^*)_{12} \\ 
(\Theta_r^*)_{21} & (\Theta_r^*)_{22} 
\end{pmatrix}.
$$

\begin{figure}[!htb]
    \centering
    \captionsetup{width=.9\linewidth}
    \includegraphics[width = 0.7\textwidth]{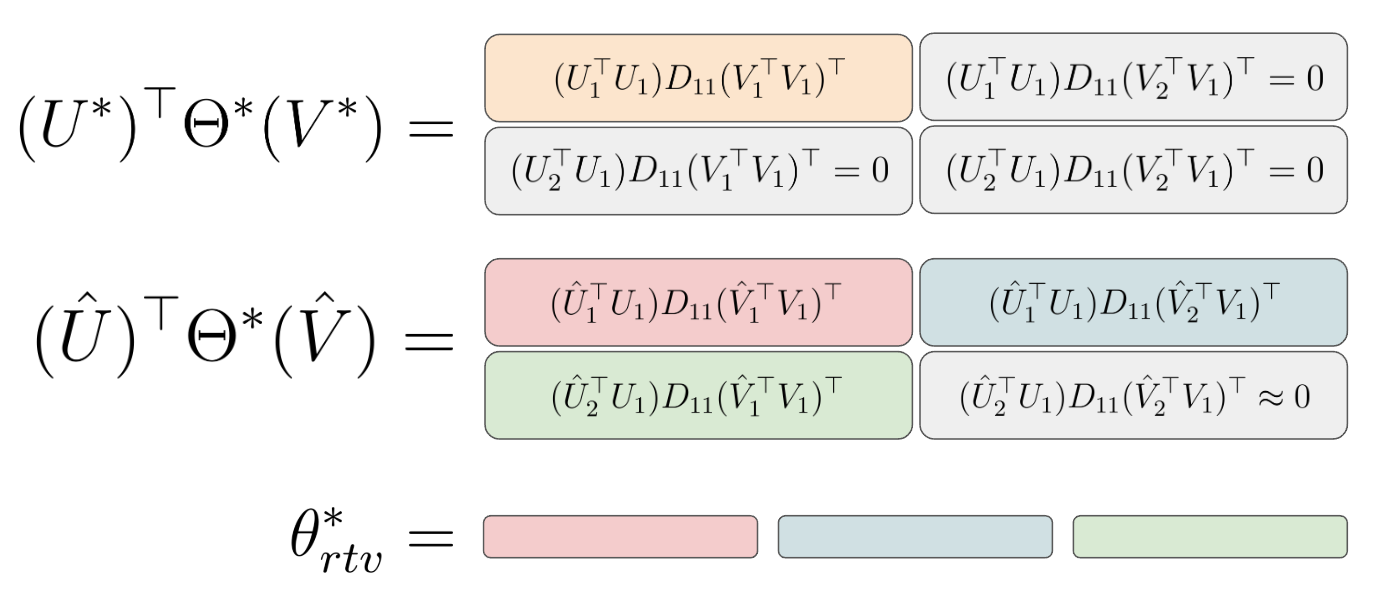}
    \caption{Illustration of rotation-truncation-vectorization.}
    \label{fig:illust_rtv}
\end{figure}

Then we employ the ``subtraction method" commonly used in the low-rank matrix bandit literature \citep{kveton2017stochastic, jun2019bilinear, kang2022efficient}, to reduce the dimension. As illustrated in Figure \ref{fig:illust_rtv}, when $\hat{U}, \hat{V}$ are close to the true subspaces, the block $(\Theta_r^*)_{22}$ is expected to be negligible as it is the product of two estimation error terms $\hat{U}_2^\top U_1$ and $\hat{V}_2^\top V_1$. By truncating this block, we retain only the essential parameters:
\begin{equation}\label{eqn:rtv}
\theta^*_{rtv} := (\vect((\Theta^*_r)_{11})^\top , \vect((\Theta^*_r)_{12})^\top , \vect((\Theta^*_r)_{21})^\top)^\top \in {\bbR}^{(d_x+d_\phi)r-r^2}.
\end{equation}

This reduced vector $\theta^*_{rtv} \in \mathbb{R}^{(d_x+d_\phi)r-r^2}$ serves as the basis for efficient uncertainty quantification. Now the contextual preference model can be reformulated as:
$$
r_{\Theta^*}(x,s,a) = x^\top \Theta^* \phi(s,a) = (\hat{U}^\top x)^\top (\Theta^*_r) (\hat{V}^\top \phi(s,a)) = \langle(\hat{U}^\top x) (\hat{V}^\top \phi(s,a))^\top , \Theta^*_r \rangle.
$$

Here, $(\hat{U}^\top x) (\hat{V}^\top \phi(s,a))^\top = \hat{U}^\top (x \phi(s,a)^\top ) \hat{V}$ is the rotation of $x\phi(s,a)^\top$ with respect to $\hat{U}$ and $\hat{V}$. Let $Z = x\phi(s,a)^\top$ and $z_{rtv}$ denote the rotation-truncation-vectorization of $Z$. Then, the contextual reward function can be expressed as:
$$
r_{\Theta^*}(x,s,a) = \langle Z, \Theta^* \rangle = \langle z_{rtv}, \theta^*_{rtv} \rangle + \langle (Z_r)_{22}, (\Theta^*_r)_{22} \rangle \approx \langle z_{rtv}, \theta^*_{rtv} \rangle,
$$
where the approximation holds as $(\Theta^*_r)_{22} \approx 0$. This reformulation allows us to focus on estimating the low-dimensional parameter $\theta^*_{rtv} \in \bbR^k$, where $k = (d_x + d_\phi)r - r^2$ rather than the full matrix $\Theta^* \in \bbR^{d_x\times d_\phi}$. This significantly reduces the computational complexity while maintaining the model's accuracy.

After reducing the problem to a low-dimensional subspace, we re-estimate the parameter in this reduced space. For the remaining $N-N_0$ observations, define $Z_t = x_t ( \phi(s_t,a_{t1}) - \phi(s_t, a_{t0}))^\top$ and let $z_{t,rtv}\in \bbR^k$ be the rotation-truncation-vectorization of $Z_t$ for $t> N_0$. The approximate negative log-likelihood function in the reduced space is defined as:
\begin{equation}\label{eqn:likelihood_rtv}
{\calL}_{rtv}(\theta_{rtv}) = \frac{1}{N-N_0} \sum_{t=N_0+1}^N \left( \log \left( 1 + \exp(-z_{t,rtv}^\top \theta_{rtv})) \right) + (1-y_t) z_{t,rtv}^\top \theta_{rtv}) \right),
\end{equation}
where $\theta_{rtv} \in \bbR^k$. As the approximate likelihood function is convex with respect to $\theta_{rtv}$,  convex optimization methods can be applied to obtain the estimate $\hat\theta_{rtv}$. 

\subsection{Pessimism in the Reduced Low-rank Space}

Finally, we apply the pessimistic approach, which is widely used in offline reinforcement learning \citep{li2022pessimism, zhu2023principled}. Instead of the greedy approach that maximizes the estimated reward function based on the estimated parameter, the pessimistic approach maximizes the \textbf{pessimistic reward}. This reward is defined by penalizing the estimated reward based on the uncertainty quantification of the parameter. Intuitively, actions are penalized if the uncertainty in the estimated reward is high. This approach mitigates the risk of selecting suboptimal actions that may result from inaccurate estimates due to the distribution of the offline dataset.

To quantify uncertainty, we define the empirical second moment in the reduced space: $W = \frac{1}{n_0} \sum_{t=1}^{n_0} z_{t,rtv}z_{t,rtv}^\top$. Using this, we construct the confidence set $\{\theta_{rtv}: \|\theta_{rtv} - \hat\theta_{rtv}\|_W \le C\}$ where $\|\cdot\|_W$ represents the Mahalanobis norm weighted by $W$, and $C$ is the confidence threshold.

To define the pessimistic reward based on this confidence set, we introduce the approximate reward function with respect to $\theta_{rtv}$:
$$
r_{\theta_{rtv}}(x,s,a) = z_{rtv}(x,s,a)^\top \theta_{rtv},
$$
where $z_{rtv}(x,s,a)$ is the rotation-truncation-vectorization of $x\phi(s,a)^\top$. The pessimistic value function is then defined as:
$$
\hat{J}(\pi) := \min_{\|\theta_{rtv} - \hat\theta_{rtv}\|_W \le C} {\bbE}_{(x,s)\sim \rho} r_{\theta_{rtv}}(x,s,\pi).
$$
The complete PRS algorithm is summarized in Algorithm \ref{alg:prs}.

\begin{algorithm}
 \caption{Pessimism in Reduced Space (PRS) Algorithm}
 \label{alg:prs}
 \begin{algorithmic}[1]
 \Require Dataset $\{(x_i, s_i, a_{i0}, a_{i1}, y_i : i\in [N]\}$, rank of parameter matrix $r$, learning rate $\eta$, initial parameter estimate $\Theta_0$ for Algorithm \ref{alg:fgd} and subspace exploration length $N_0$.
 \State Obtain the rank-constrained MLE $\hat\Theta = \hat{U} \hat{D} \hat{V}^\top$ by applying Algorithm \ref{alg:fgd} to the first $N_0$ observations;
 \State Rotate-truncate-vectorize the remaining $N-N_0$ observations and find the maximizer of (\ref{eqn:likelihood_rtv});
 \State Construct confidence set $\mathbf{\Theta}_B = \{\theta_{rtv}: \|\theta_{rtv} - \hat\theta_{rtv}\|_W \le C\}$;
 \State Construct the pessimistic value function $\hat{J}(\pi) = \min_{\theta_{rtv} \in \mathbf{\Theta}_B} \bbE_{(x,s) \sim \rho}  r_{\theta_{rtv}}(x,s, \pi)$;
 \State Calculate $\hat\pi = \argmax_{\pi} \hat{J}(\pi)$.
 \end{algorithmic}
\Return The pessimistic policy $\hat\pi$.
\end{algorithm}

In case where the policy can be parametrized as $\pi_\theta$, finding the pessimistic policy can be solved via gradient based methods.

\section{Theory}

In this section, we first present the assumptions required to establish the rate of the sub-optimality gap. We then derive the estimation error for the low-rank subspace estimation and provide an upper bound on the sub-optimality gap for the PRS algorithm.

\subsection{Assumptions}

We begin by introducing the boundedness condition and the positive definite second moment condition of the features and the parameter.

\begin{assm}[Boundedness]\label{assm:bounded_features}
    The expert feature $x_t$ is bounded by $B_x$, i.e., $\|x_t\|_2 \le B_x$. For any query $s\in \calS$ and answer $a\in \calA$, the feature embedding $\phi$ is bounded, i.e.,  $\|\phi(s,a)\|_2 \le B_{\phi}$. For the matrix parameter $\Theta^*$,  $\|\Theta\|_F \le B_{\theta} $ for all $\Theta \in \mathcal{B}(\Theta^*,1)$.
\end{assm}

\begin{assm}[Positive Definite Second Moment]\label{assm:cov}
Suppose the context-state-actions $(x_t, s_t, a_{t1}, a_{t2}) \sim P$ for the data distribution $P$. Let $z_t = (\phi(s_t, a_{t1}) - \phi(s_t, a_{t0})) \otimes x_t$. Then $\bbE_P z_t z_t^\top = \Sigma \succ 0$. 
\end{assm}

The boundedness assumption is common in contextual reinforcement learning problems \citep{li2017provably, zhan2023provable}. The invertibility of the second moment of the feature is required to guarantee that the parameter is identifiable. Suppose $\bbE_P x_t x_t^\top = \Sigma_x \succ 0$ and $\bbE_P(\phi(s_t, a_{t1}) - \phi(s_t, a_{t0}))(\phi(s_t, a_{t1}) - \phi(s_t, a_{t0}))^\top = \Sigma_\phi \succ 0$, with $x_t \perp (s_t, a_{t0}, a_{t1})$. Then $\Sigma = \Sigma\phi \otimes \Sigma_x$ and therefore $\Sigma \succ 0$. In other words, if the features from the state-action pairs and contexts both have positive definite second moment and are independent, Assumption \ref{assm:cov} is satisfied. When $x_t$ is uniformly distributed over standard unit vectors of $\bbR^M$, it becomes equivalent with the problem in \citet{zhong2024provable}, which assumes each query-answer pair is given to each of the $M$ user types.

\begin{rmk}[Interpretation of Theoretical Assumptions in RLHF]
The boundedness assumption (Assumption \ref{assm:bounded_features}) is standard in preference-based RLHF \citep{zhu2023principled, zhong2024provable}. Under the BTL model, preference probabilities depend exponentially on reward differences. If feature magnitudes or reward values are allowed to grow without control, moderate differences already lead to near-deterministic comparisons. In such regimes, observed feedback carries vanishing information, making offline estimation statistically infeasible. In practice, boundedness can be enforced through normalization or clipping of learned representations. Moreover, the positive definiteness assumption (Assumption \ref{assm:cov}) on the second-moment matrix ensures identifiability of the reward parameters under preference feedback and is also standard in preference learning and RLHF \citep{zhong2024provable}. This condition rules out degenerate or redundant feature directions and guarantees that each parameter dimension is identifiable from the observed comparisons. 
\end{rmk}

\subsection{Main Results}

We first present the estimation bound of the low-rank estimator.

\begin{thm}[Informal Estimation Bound] \label{thm:estimation_bound_informal}
    Suppose Assumptions \ref{assm:bounded_features} and \ref{assm:cov} hold. Let $r$ denote the rank of $\Theta^*$, with $\sigma_1 \ge \sigma_2 \ge \cdots \sigma_r > 0$ as its singular values. Assume we have an initial estimate $\Theta_0 \in \bbR^{d_x\times d_\phi}$ such that $\|\Theta_0 - \Theta^*\|_F \le c_2 \sqrt{\sigma_r}$ for some constant $c_2$. Then for the low-rank estimator $\hat\Theta$ achieved from Algorithm \ref{alg:fgd} with learning rate $\eta = c_1/\sigma_1$, we have the estimation bound:
    $$
    \|\hat\Theta - \Theta^*\|_F \le C_1 \cdot  \sqrt{\frac{\sigma_1 r\log(\frac{d_x + d_\phi}{\delta})}{\sigma_r N_0}}
    $$
    with probability at least $1-\delta$ for some constant $C_1 > 0$.
\end{thm}

\begin{rmk} [Initial Estimate $\Theta_0$]
    We assume the availability of an initial estimate $\Theta_0$ that is reasonably close to $\Theta^*$. This assumption is standard in low-rank matrix estimation problems \citep{jain2013low, chi2019nonconvex, xia2021statistical} and is commonly referred to as the ``basin of attraction" condition, ensuring that the algorithm converges to the desired target. This assumption holds in various scenarios. For instance, when the sample size is sufficiently large, the unconstrained maximum likelihood estimator can serve as a warm start for Algorithm \ref{alg:fgd}, providing an initial estimate close to $\Theta^*$.
\end{rmk}

The estimation error bound is proportional to $\sqrt{r}$, whereas without the low-rank assumption, it would scale with $\sqrt{\min\{d_x, d_\phi\}}$. This highlights the critical role of the low-rank assumption in reducing estimation error. This rate aligns with bounds in the low-rank matrix estimation literature, particularly under the assumption of bounded $\ell_2$-norms for the context $x$ and feature embedding $\phi(s,a)$. Specifically, if the max-norm of the parameter matrix is bounded, the $\ell_2$-norm bound scales with $d_x$ and $d_\phi$, consistent with results in \citet{negahban2011estimation}, \citet{xia2021statistical}, and \citet{zhu2022learning}. Building on this result, we derive the following corollary concerning the error induced by the rotation-truncation-vectorization process.

\begin{cor} [Error induced by Rotation-Truncation-Vectorization]
    Suppose $\|\hat\Theta - \Theta^*\|_F \le B$ for some $B > 0$. Then,
    \begin{equation} \label{eqn:S_perp}
    \|(\Theta^*_r)_{22}\|_F \le  C_2 \cdot  \frac{\sigma_1}{\sigma_r^2} \cdot B^2
    \end{equation}
    for some constant $C_2 > 0$.
\end{cor}

The details of the proof are provided in Appendices \ref{sect:EstBound} and \ref{sect:rtv_error}. Using these results, we establish the confidence bounds for the estimator $\hat\theta_{rtv}$ in the reduced space.

\begin{lem}[Confidence Bound on $\theta_{rtv}$]
\label{lem:confidence_bound}
    Suppose Assumptions \ref{assm:bounded_features} and \ref{assm:cov} hold. Further assume that $\|(\Theta_r^*)_{22}\|_F \le S_\perp$ for some $S_\perp$. Let $\hat\theta_{rtv}$ be the minimizer of the approximate negative log-likelihood function (\ref{eqn:likelihood_rtv}). Then,
    \begin{align*}
    \|\hat{\theta}_{rtv} - \theta^*_{rtv}\|_W
    &\le \frac{1}{\gamma} \left( \sqrt{C_3 \cdot \frac{(d_x+d_\phi)r + \log(1/\delta)}{N-N_0}} +  \frac{1}{2} B_x B_\phi \cdot S_\perp \cdot \sqrt{(d_x+d_\phi)r} \right) 
    \end{align*}
    with probability at least $1-\delta$.
\end{lem}

Finally, we derive an upper bound on the sub-optimality gap for the proposed policy.

\begin{thm}[Upper Bound of Sub-optimality] \label{thm:subopt}
    Suppose Assumptions \ref{assm:bounded_features} - \ref{assm:cov} hold with $n_0 = \Omega(n)$ and $n_1 = n-n_0 = \Omega(n)$. Then,
    $$
    J(\pi^*) - J(\hat\pi) = O\left( C^* \cdot \sqrt{\frac{(d_x+d_\phi)\cdot r + \log(1/\delta)}{N}}\right)
    $$
    with probability at least $1-\delta$, where $C^* = \left\|  {\bbE}_{(x,s)\sim\rho} z_{rtv}(x,s,\pi^*) \right\|_{W^{-1}}$.
\end{thm}

In low-rank case where $r \ll \min\{d_x, d_\phi\}$, our bound represents a significant improvement over $O\Big(C^* \cdot \sqrt{\frac{d_xd_\phi + \log(1/\delta)}{n}}\Big)$ achieved by existing methods \citep{zhu2023principled}. Moreover, the setting in \citet{zhong2024provable} is a special case of our framework when the true clustering structure of human feedback is known with $M = d_x$ and $d =d_\phi$. They derived a sub-optimality rate of 
    $$
    M \sqrt{r^2/n + (dr^2 \log{d} + r\log(M/\delta))/(Mn)} \cdot \tilde{C}^*,
    $$ 
    where $\tilde{C}^* = \max_m \|(\Sigma_m^{-1/2} \bbE_{s\sim \rho} \phi(s, \pi^*(s))\|_2$. Assume $r \ll M \ll d$, and ignoring logarithmic factors, their rate simplifies to $M \cdot \sqrt{\frac{(M+d)r^2 + r\log(\frac{M}{\delta})}{Mn}} \cdot \tilde{C}^*$. Noting that their value function is defined as the sum over $M$ types for the utilitarian welfare function, and we can verify that our rate of $\tilde{C}^* \cdot \sqrt{\frac{(M+d)r + \log(\frac{1}{\delta})}{n}}$ matches this bound in this special case.

\begin{rmk} [Discussion on Concentratability Coefficient]
The concentratability coefficient $C^*$ is invariant under rotation and decreases as covariates are removed. Since the rotation in the bilinear form remains a rotation after vectorization, the rotation-truncation-vectorization (RTV) process reduces the concentratability coefficient. When the matrix is full-rank, RTV simplifies to a pure rotation, ensuring consistency in the concentratability coefficient. Furthermore, this aligns with the concentratability coefficient defined in \citet{zhu2023principled} when $x = 1$, i.e., in the absence of contextual information. The theoretical details are provided in Appendix \ref{sect:ConcenCoeff}.
\end{rmk}

The analysis of our algorithm presents several challenges. First, the rank-constrained maximum likelihood estimation problem is inherently non-convex, which precludes the direct application of classical results from the convex optimization literature. Moreover, the offline dataset comprises binary responses from human experts, rather than quantitative data. This characteristic complicates the derivation of the estimation bounds, as it requires handling the discrete nature of the data. During the uncertainty quantification step, both the estimation error of the subspace and the likelihood estimation contribute to the overall uncertainty of the parameter estimate in the reduced space. Lastly, the approximation of the preference function in the reduced space is similarly influenced by the subspace estimation error, which propagates through the analysis and affects the policy's value. To address these challenges, we develop novel tools that incorporate subspace estimation error into the construction of estimation bounds and the approximation of the preference function. These tools provide insights into how the subspace estimation error impacts the value of the policy, ensuring a rigorous analysis of our algorithm.

\section{Numerical Studies}

\subsection{Simulation Studies}

In this section, we compare the sub-optimality gap of the proposed policy with several benchmarks in the simulations.
We set $\calX = \bbR^{d_x}$ and $\calS = \bbR^{d_s}$, and generate individual user contexts $x_t$ and states $s_t$ independently from $\mathrm{Unif}([-1,1])$.
The action space is $\calA = \{0,1,\ldots,d_a-1\}$. 

To generate the state-action feature representation $\phi(s,a)$, 
we first generate a matrix $M \in \bbR^{d_s \times d_s}$ by sampling each entry independently from $\mathrm{Unif}([-1,1])$ and normalizing it to satisfy $\|M\|_F = 1$.
The feature mapping $\phi : \calS \times \calA \to \bbR^{d_\phi}$, where $d_\phi = d_a + d_s-1$, is defined as
$$
\phi(s,a) =
\begin{cases}
(0_{d_a-1}^\top,\; 0.1\, M s^\top), & a = 0, \\
(e_a^\top,\; 0.1\, M s^\top), & a \ge 1,
\end{cases}
$$
where $0_{d_a-1} \in \bbR^{d_a-1}$ is the zero vector and $e_a \in \bbR^{d_a-1}$ is the $a$-th standard basis vector.
This construction applies one-hot encoding to the categorical action variable and concatenates it with a scaled state-dependent feature.

The reward parameter matrix $\Theta\in\bbR^{d_x\times d_\phi}$ is constructed to have rank $r$. For each $k\in{1,\ldots,d_a-1}$, define

$$
w_k = \left( \frac{1}{k}, \frac{1}{k+1}, \dots, \frac{1}{k+r-1}, 0, \dots, 0 \right)^\top \in {\bbR}^{d_x},
$$
so that only the first $r$ components are non-zero, and let $v_k = w_k / \|w_k\|_2$. We then set $\Theta_{:,k} = 2\cdot v_k$, for $k = 1,\ldots,d_a-1,$
with all remaining columns corresponding to state features set to zero. The scaling factor $2$ controls the signal strength. By construction, $\Theta$ has rank $r$, inducing a smoothly varying low-rank structure across actions. 

To model heterogeneous coverage in offline preference data, we introduce an imbalanced distribution over action pairs controlled by a parameter $q$. Specifically, for a fraction $1/q$ of the samples, action pairs are drawn uniformly from $\{(0,a_1) \in \calA\times \calA : a_1 \ne 0\}$, while for the remaining fraction $1-1/q$, the action pair is fixed as $(0,1)$. As $q$ increases, the majority of comparisons involve only actions $0$ and $1$, resulting in increasingly imbalanced feedback

We generate parameters, contexts, and features to simulate human feedback and assess the policy on a new set of contexts $x$ and states $s$. For comparison, we use two baseline policies.

\begin{itemize}
    \item \textbf{MLE-Greedy}: This policy uses the MLE obtained from the unconstrained maximum likelihood problem. Since the negative log-likelihood function is convex, the estimator can be efficiently computed. Using the MLE $\hat\Theta = \argmax_\Theta \calL(\Theta)$ with $\calL$ defined in (\ref{eqn:likelihood}), the policy evaluates the estimated value of each action and selects the greedy action:
    $$
    \hat\pi_{greedy}(x,s) = \argmax_{a\in \calA} r_{\hat\Theta}(x,s,a).
    $$
    \item \textbf{MLE-Pessimistic}: 
    Similar to MLE-Greedy, this policy uses the MLE $\hat{\Theta}$ but incorporates the pessimistic approach from \citet{zhu2023principled}. To adapt the bilinear formulation of the reward to the linear framework used in \citet{zhu2023principled}, the feature vector is modified to $z_t = \phi(s_t, a_t) \otimes x_t \in \mathbb{R}^{d_x d_\phi}$, representing the vectorization of the outer product. This feature vector is used to quantify the confidence bound. Given the pessimistic value function
    $
    \hat{J}(\pi) = \min_{\Theta \in CB(\hat\Theta)} {\bbE}_{(x,s)\sim \rho} r_{\Theta}(x,s,a) 
    $
   with  $CB(\hat\Theta)$ being the confidence bound on the vectorized feature, this policy maximizes the pessimistic value function:
    $$
    \hat\pi_{\text{pess}} = \argmax_\pi \hat{J}(\pi).
    $$
\end{itemize}

We evaluate our proposed PRS policy alongside the MLE-Greedy and MLE-Pessimistic policies as baselines. For details of implementing the policy optimization problem $\hat\pi = \argmax \hat{J}(\pi)$, we refer to Appendix \ref{sect:PolicyOpt}. For PRS and MLE-Pessimistic, we use the approximate solution defined in (\ref{eqn:approximate_policy}) of the Appendix. Each simulation is repeated 20 times, and the sub-optimality gap is reported for $N\in \{1000, 2000, 3000, 5000, 10000\}$ under various scenarios.

\begin{figure}[!htb]
    \centering
    \captionsetup{width=.9\linewidth}
    \includegraphics[width = \textwidth]{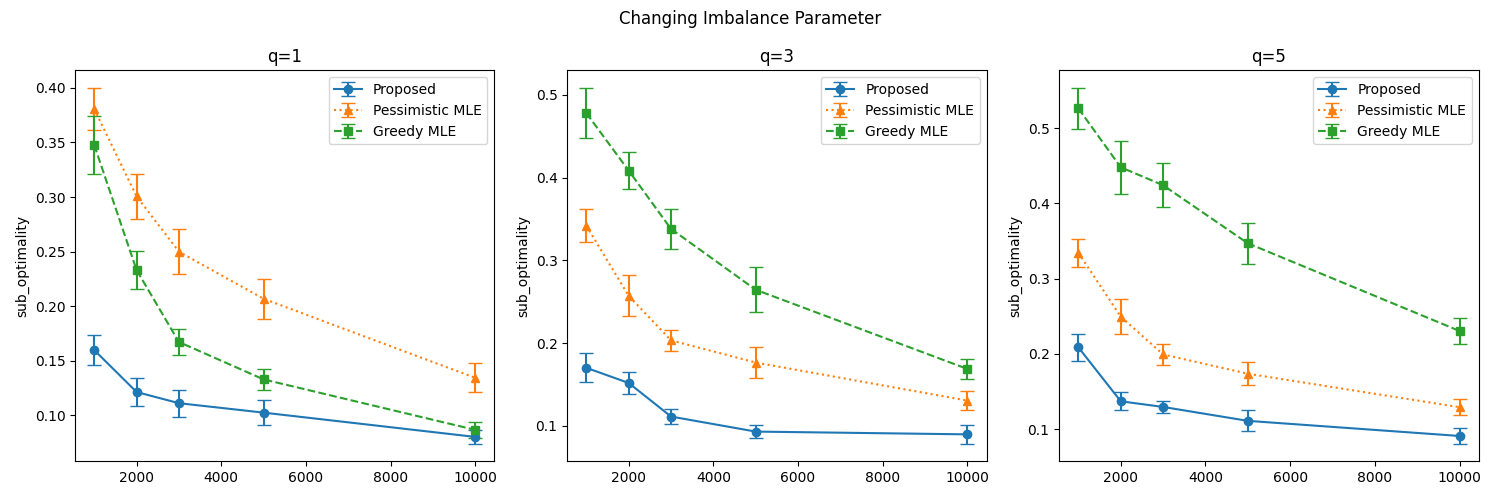}
    \caption{Sub optimality gap of our proposed PRS policy compared with MLE-Greedy and MLE-Pessimistic when the imbalance differs in datasets. }
    \label{fig:fig_change_a_sparsity}
\end{figure}

We first study the impact of offline data imbalance on policy performance.
We set $d_s = 20, d_a = 10, d_x = 20$ with rank $r=3$, resulting in $\Theta \in \bbR^{20\times 29}$. As shown in Figure\ref{fig:fig_change_a_sparsity}, as the imbalance parameter $q$ increases, all methods exhibit larger sub-optimality gaps. Among the policies, MLE-Greedy consistently performs the worst, while the proposed PRS policy achieves the best performance across all settings. Moreover, the gap between the methods widens as the data imbalance increases, highlighting the robustness of the PRS policy to variations in offline data distribution.

\begin{figure}[!htb]
    \centering
    \captionsetup{width=.9\linewidth}
    \includegraphics[width = \textwidth]{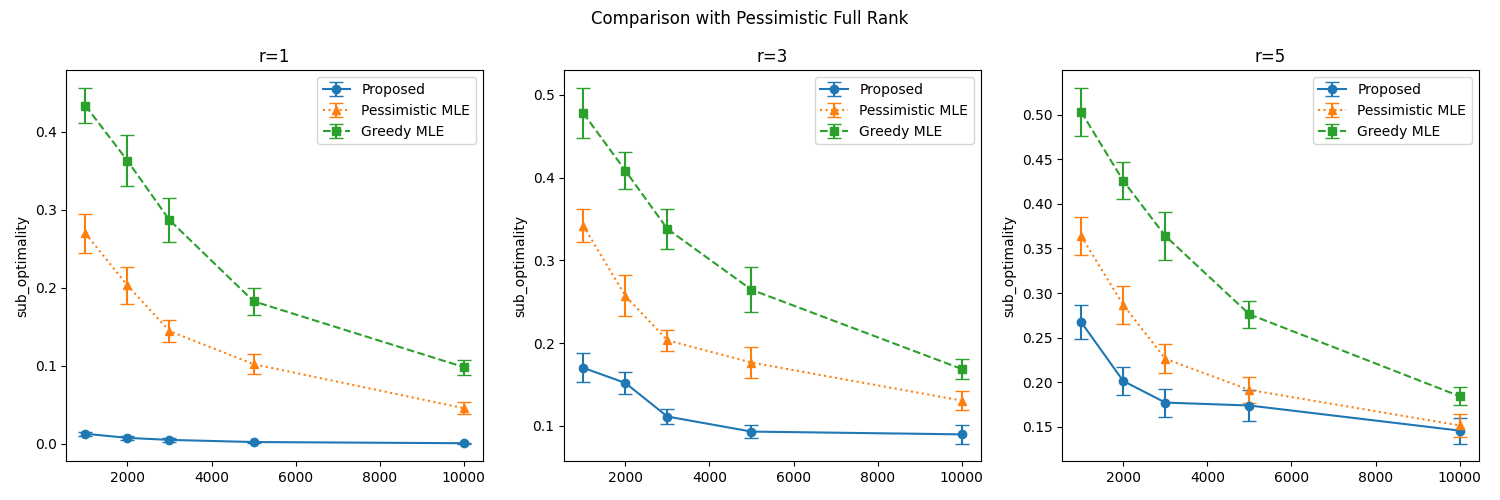}
    \caption{Sub optimality gap of our proposed PRS policy compared with MLE-Greedy and MLE-Pessimistic when the true rank differs.}
    \label{fig:fig_change_r}
\end{figure}

Next we evaluate the effect of the low-rank assumption on policy performance. Similar to the first experiment, we use $d_s = 20, d_a = 10, d_x = 20$, but here we fix imbalance parameter $q=3$ and vary the ranks $r\in \{1,3,5\}$. As shown in Figure \ref{fig:fig_change_r}, across all ranks, the proposed PRS method consistently outperforms the MLE-Greedy and MLE-Pessimistic policies. The sub-optimality gap is nearly zero when $r=1$, indicating excellent performance under strong low-rank assumptions. As the rank increases, the performance gap narrows, indicating diminishing gains as the reward structure becomes less compressible.

\subsection{PersonalLLM Dataset}

\begin{figure}[!htb]
    \centering
    \captionsetup{width=.9\linewidth}
    \includegraphics[width=0.9\textwidth]{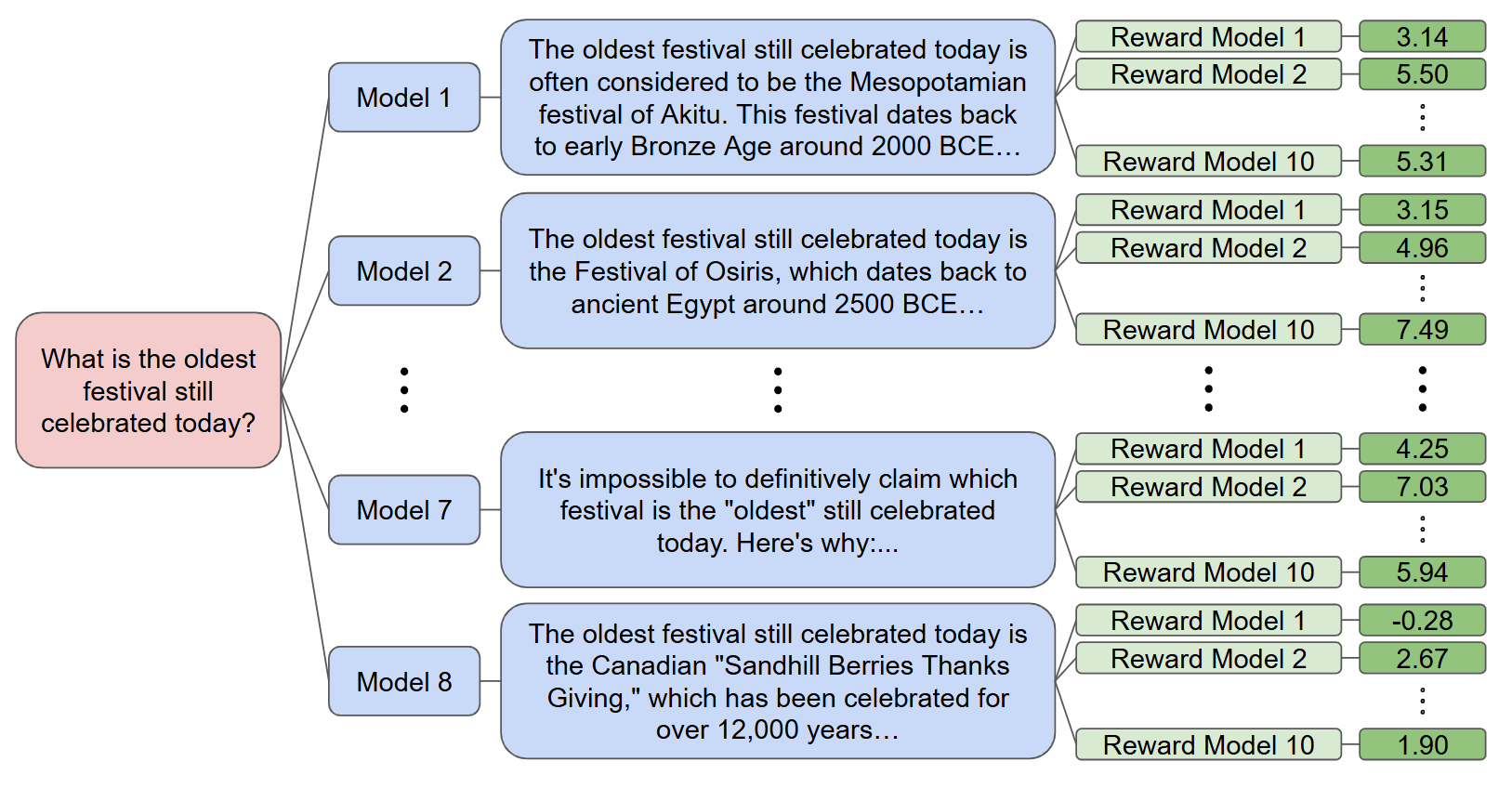}
    \caption{Illustration of the PersonalLLM dataset. For each prompt, multiple responses generated by different LLMs are evaluated by a collection of reward models.}
    \label{fig:illust_personalllm}
\end{figure}

We next evaluate our method on the PersonalLLM dataset \citep{zollo2024personalllm}, a publicly released benchmark designed to study personalization in large language models.
The dataset consists of prompts covering a wide range of topics and intents (9{,}402 for training and 1{,}000 for testing), each paired with responses generated by 8 different large language models, including GPT-4o, Claude~3, Llama~3, and Gemini~Pro~1.5.
Each response is further evaluated by 10 different reward models, yielding heterogeneous preference signals, as illustrated in Figure~\ref{fig:illust_personalllm}.

Following the experimental procedure of \citet{bose2025lore}, we generate user contexts from a uniform distribution.
For each prompt-response pair, we collect the scores produced by the 10 reward models and standardize them to have mean $0$ and unit variance. These standardized scores are used as the feature representation $\phi(s,a)$, where $s$ denotes the prompt and $a$ denotes the candidate response. 
We then construct a low-rank reward parameter matrix using the same procedure as in the simulations.

\begin{figure}[!htb]
    \centering
    \captionsetup{width=.9\linewidth}
    \includegraphics[width=0.9\textwidth]{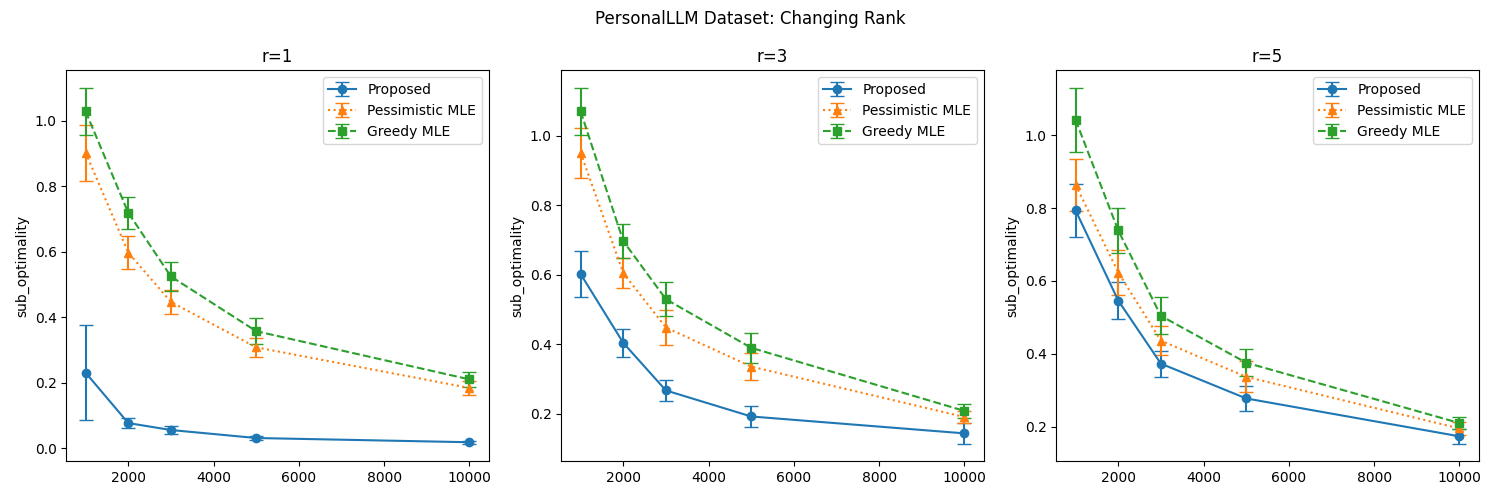}
    \caption{Sub-optimality gap of the proposed PRS policy compared with MLE-Greedy and MLE-Pessimistic with varying ranks on the PersonalLLM dataset.}
    \label{fig:fig_personalllm}
\end{figure}

We compare the proposed PRS policy with the MLE-Greedy and MLE-Pessimistic baselines.
In the training data, for each prompt, we randomly sample a pair of responses from the 8 candidate LLM outputs and generate a binary preference label according to the BTL model.
We then evaluate the sub-optimality gap of each policy relative to the optimal decision rule on the test dataset.
As shown in Figure \ref{fig:fig_personalllm}, the proposed PRS policy achieves consistently lower sub-optimality gaps than the baseline methods for different ranks, illustrating the benefit of incorporating low-rank contextual structure in preference-based RLHF problems.

\begin{figure}[!htb]
    \centering
    \captionsetup{width=.9\linewidth}
    \includegraphics[width=0.9\textwidth]{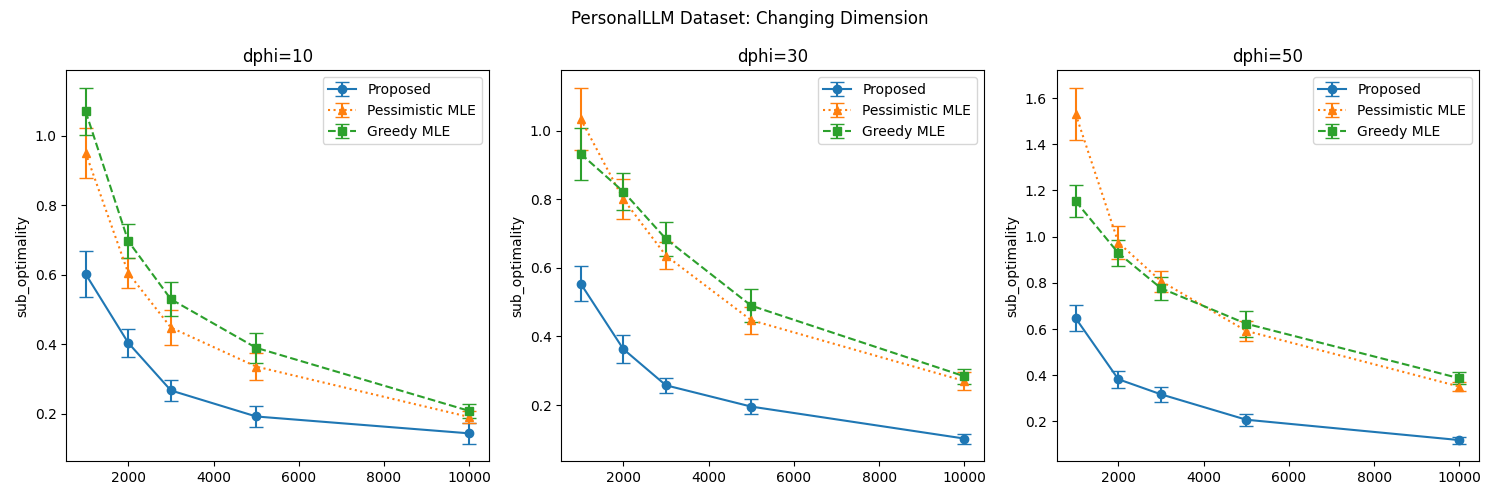}
    \caption{Sub-optimality gap on the PersonalLLM dataset with augmented noisy features.}
    \label{fig:fig_personalllm_dim}
\end{figure}

To further assess robustness, we evaluate performance under the presence of irrelevant noisy features.
Specifically, we augment $\phi(s,a)$ by concatenating additional noise dimensions, thereby increasing the ambient feature dimension without changing the underlying preference structure.
As shown in Figure~\ref{fig:fig_personalllm_dim}, the proposed PRS policy maintains stable performance, whereas the baseline methods deteriorate as the feature dimension increases.

\baselineskip=22pt
\setstretch{1.35}
\bibliography{ref}

\appendix
\doublespacing

\newpage
\appendix 
\baselineskip=24pt
\setcounter{page}{1}
\setcounter{equation}{0}
\setcounter{section}{0}
\renewcommand{\thesection}{S.\arabic{section}}
\renewcommand{\theequation}{S\arabic{equation}}

\begin{center}
{\Large\bf Supplementary Materials} \\
\medskip
{\Large\bf ``Low-Rank Contextual Reinforcement Learning from Heterogeneous Human Feedback"}  \\
\bigskip
\end{center}
\bigskip

\noindent

This supplement provides additional empirical results, implementation details, additional discussions, and all technical proofs. In particular, Section \ref{sec:rank_estimation} studies the rank estimation of the proposed low-rank model. Section \ref{sec:sensitivity} presents sensitivity analyses, including robustness to data splitting and to the choice of the pessimism hyperparameter in the confidence bound. Section \ref{sec:runtime} reports runtime results, and Section \ref{sec:discussion} discusses several promising future research directions. Section \ref{sect:PolicyOpt} details the implementation of pessimistic policy optimization. Finally, Section \ref{sect:AppendixProof} collects the technical proofs of all supporting lemmas and theorems.

\section{Rank Estimation}\label{sec:rank_estimation}

\begin{figure}[!htb]
    \centering
    \captionsetup{width=.9\linewidth}
    \includegraphics[width = 0.6\textwidth]{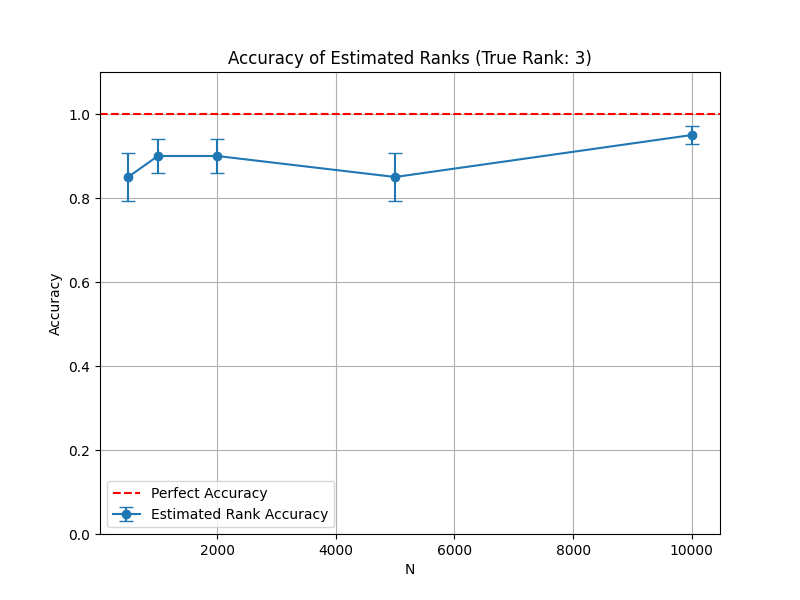}
    \caption{Accuracy of rank estimation using GIC.}
    \label{fig:fig_rank_est}
\end{figure}

In this section, we study the empirical performance of rank selection for the proposed low-rank model. Following standard practice, we adopt a generalized information criterion (GIC) \citep{konishi1996generalised, fan2013tuning} to select the rank by balancing goodness of fit and model complexity. Specifically, for each candidate rank $r$, we compute the penalized objective defined as:
$$
GIC(r) = \mathcal{L}_n(\hat{\Theta}_r) + a_n \cdot r(d_x+d_\phi - r),
$$

where $\mathcal{L}_n$ denotes the negative log-likelihood, $\hat{\Theta}_r$ is the rank-$r$ constrained minimizer of $\mathcal{L}_n$ and $a_n$ is the penalty parameter. Following \cite{lee2024low}, we set $a_n = C \cdot \log{n}/n$ with $C = 0.1$ to ensure consistency of rank selection. We estimate the rank estimation accuracy under the setting of our simulation studies in Section 5.1, with $d_s = 20, d_a = 10, d_x = 20$ with rank $r=3$ and imbalanced action distribution with parameter $3$ and a diagonal matrix with diagonal entries $2$. Figure \ref{fig:fig_rank_est} reports the accuracy of the estimated rank as a function of the sample size $N$. The results show that the proposed procedure reliably recovers the true rank as the sample size increases.

\section{Sensitivity Analysis}
\label{sec:sensitivity}

We conduct additional sensitivity analyses to assess the robustness of the proposed method to data splitting and to the choice of the pessimism parameter $C_2$.

\subsection{Data Split}

We examine the sensitivity of the proposed PRS policy to different random splits of the offline dataset into training and evaluation sets. We use $N/N_0 \in \{0.2, 0.5, 0.8\}$ and a ``No Split" setting where the entire dataset is used both for subspace estimation and uncertainty quantification in the reduced space. Similar to our simulation studies in Section 5.1, we use $d_s = 20, d_a = 10, d_x = 20$ with rank $r=3$ and imbalance parameter $3$.

\begin{figure}[h!]
    \centering
    \captionsetup{width=.9\linewidth}
    \includegraphics[width = 0.6\textwidth]{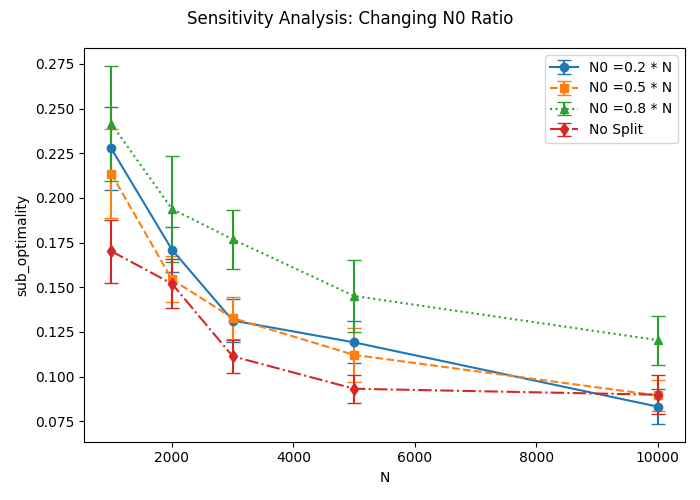}
    \caption{Sub-optimality of proposed method under different data splits.}
    \label{fig:fig_change_N0}
\end{figure}

As shown in Figure \ref{fig:fig_change_N0}, the performance of the algorithm is robust to the choice of the data split ratio provided that $N_0 = \Omega(N)$. While our theoretical analysis requires $N_0$ and $N$ to be of the same order, empirical results indicate that performance is strongest in the no-split setting. Motivated by this observation, we adopt the no-split configuration in all subsequent numerical experiments.

\subsection{Confidence Bound Hyperparameter}

\begin{figure}[!htb]
    \centering
    \captionsetup{width=.9\linewidth}
    \includegraphics[width = 0.6\textwidth]{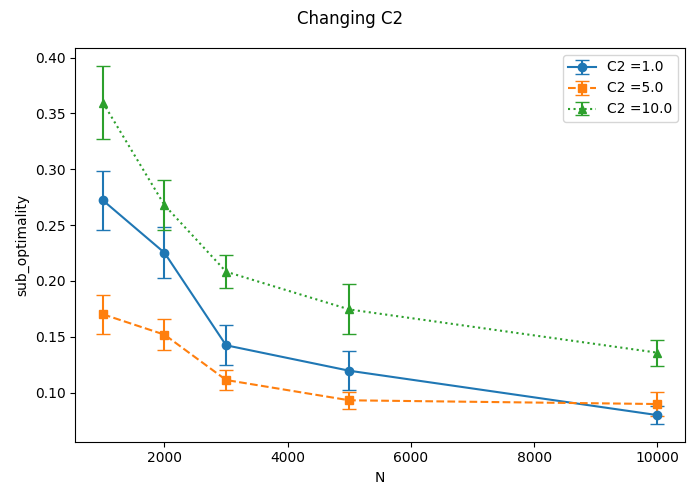}
    \caption{Sub-optimality of proposed method under different confidence bound parameter $C_2$.}
    \label{fig:fig_change_C2}
\end{figure}

We further conduct a sensitivity analysis on hyperparameters of the confidence bound. The confidence threshold $C$ is our PRS algorithm is derived from Lemma \ref{lem:confidence_bound}. For practical implementation, we use the formula $C = C_2 \cdot \sqrt{\frac{(d_x+d_\phi)r}{N-N_0}}$, where $C_2$ is a hyperparameter. We analyze the sub-optimality of our proposed method over $C_2 \in \{1,5,10\}$ as shown in Figure \ref{fig:fig_change_C2}. When $C_2$ is too small, the resulting confidence region is overly narrow and the algorithm derives limited benefit from pessimism; when $C_2$ is too large, excessive pessimism leads to conservative decisions and degraded performance. Based on this empirical trade-off, we set $C_2=5$ in all numerical experiments. Importantly, the choice of $C_2$ affects only the finite-sample behavior of the algorithm and does not alter the theoretical guarantees, which hold up to universal constant factors.

\section{Runtime Analysis}\label{sec:runtime}

In this section, we report the total runtime of the experiment pipeline in our simulation studies for representative feature dimensions$(d_x, d_\phi)$, with the sample size fixed at $N = 10,000$ and rank $r = 3$. The reported runtime includes data preprocessing, model initialization, and execution of all compared methods. As summarized in Table~\ref{tb:runtime}, the overall runtime increases smoothly with the feature dimensions and exhibits an approximately linear dependence on both $d_x$ and $d_\phi$. These results indicate that the proposed approach remains computationally feasible and scales favorably to moderately high-dimensional settings.

\begin{table}[h]
\centering
\begin{tabular}{ccccc}
\hline
$d_x$ & $d_\phi$ & Time (s) \\ \hline
30 & 30 & $\approx$ 51.9 \\
30 & 50 & $\approx$ 83.6 \\
50 & 50 & $\approx$ 152.7 \\
100 & 30 & $\approx$ 179.0 \\
100 & 50 & $\approx$ 251.3 \\ \hline
\end{tabular}
\caption{Average runtime across varying feature dimensions $d_x$ and $d_\phi$.}
\label{tb:runtime}
\end{table}

\section{Future Directions}\label{sec:discussion}

In this section, we highlight several potential directions for future research. In this work, we evaluate the value of a policy using the expectation of the preference function over a given distribution. This approach can be extended to optimize more complex social welfare functions, such as the Nash welfare function or the Leximin welfare function \citep{zhong2024provable}. Another promising direction is the selection of experts and queries, commonly referred to as active learning \citep{liu2024dual}. Furthermore, investigating robustness under model misspecification, potentially building on recent frameworks for covariate-assisted ranking \citep{fan2024uncertainty, fan2024covariate}, represents an important step toward ensuring reliability in complex feedback environments. Beyond addressing the heterogeneity of contexts, it is crucial to consider non-stationary environments, where the underlying distributions change over time \citep{bian2024off}. These directions offer exciting opportunities to enhance the robustness and applicability of algorithms in reinforcement learning with human feedback.

\color{black}

\section{Implementation of Policy Optimization}\label{sect:PolicyOpt}

In this section, we provide details on how to perform the policy optimization on the pessimistic value function. Note that the pessimistic value function can be simplified:
\begin{align*}
\min_{\|\theta_{rtv} - \hat\theta_{rtv}\|_W \le C} {\bbE}_{(x,s)\sim \rho} r_{\theta_{rtv}}(x,s,a) &= \min_{\|\theta_{rtv} - \hat\theta_{rtv}\|_W\le C} {\bbE}_{(x,s)\sim \rho} \theta_{rtv}^\top z_{rtv}(x,s,a) \\
&= \hat\theta_{rtv}^\top \left[{\bbE}_{(x,s)\sim \rho} z_{rtv}(x,s,a)\right] - \min_{\|\Delta\|_W \le C} \Delta^\top  \left[{\bbE}_{(x,s)\sim \rho} z_{rtv}(x,s,a)\right]\\
&\ge  \hat\theta_{rtv}^\top \left[{\bbE}_{(x,s)\sim \rho} z_{rtv}(x,s,a)\right] - \min_{\|\Delta\|_W \le C} \|\Delta\|_{W} \cdot \left\|{\bbE}_{(x,s)\sim \rho} z_{rtv}(x,s,a)\right]\|_{W^{-1}} \\
&\ge \hat\theta_{rtv}^\top \left[{\bbE}_{(x,s)\sim \rho} z_{rtv}(x,s,a)\right] - C \cdot \left\|{\bbE}_{(x,s)\sim \rho} z_{rtv}(x,s,a)\right\|_{W^{-1}}.
\end{align*}

Therefore the pessimistic policy will be
$$
\hat\pi = \argmax_\pi \left( \hat\theta_{rtv}^\top \left[{\bbE}_{(x,s)\sim \rho} z_{rtv}(x,s, \pi)\right] - C \cdot \left\|{\bbE}_{(x,s)\sim \rho} z_{rtv}(x,s, \pi)\right\|_{W^{-1}} \right).
$$

We can approximate the pessimistic policy by optimizing the relaxed pessimistic reward:
\begin{equation}\label{eqn:relaxed_reward}
\tilde\pi = \argmax_\pi \left( \hat\theta_{rtv}^\top \left[{\bbE}_{(x,s)\sim \rho} z_{rtv}(x,s, \pi)\right] - C \cdot {\bbE}_{(x,s)\sim \rho} \left[\left\|z_{rtv}(x,s, \pi)\right\|_{W^{-1}} \right] \right).
\end{equation}
Note that the pessimistic term is larger by Jensen's inequality as we now take the expectation of the norm, instead of the norm of the expectation. Then the approximate policy becomes equivalent to optimizing per $(x,s)$ pair:
\begin{equation}\label{eqn:approximate_policy}
\tilde{\pi}(x,s) = \argmax_{a\in \calA} \left( \hat\theta_{rtv}^\top z_{rtv}(x,s, a) - C \cdot \left\|z_{rtv}(x,s, a)\right\|_{W^{-1}} \right)
\end{equation}
or in the distribution shift framework optimizing with respect to the conditional distribution $x|s$:
$$
\tilde{\pi}(s) = \argmax_{a\in \calA} \left( {\bbE}_{x\sim \rho|s} \left[ \hat\theta_{rtv}^\top z_{rtv}(x,s, a) - C \cdot \left\|z_{rtv}(x,s, a)\right\|_{W^{-1}} \right] \right).
$$
This is similar to the pessimistic value iteration (PEVI) proposed by \citet{jin2021pessimism}. 

If we use the relaxed reward (\ref{eqn:relaxed_reward}) for optimizing the policy, as discussed in \citet{li2022pessimism}, the sub-optimality gap bound will also be loosened to the expectation of the norm, instead of norm of the expectation. In other words, the concentratability coefficient $C^*$ will be $\bbE_{(x,s)\sim \rho} \| z_{rtv}(x,s,\pi^*)\|_{W^{-1}}$ which will provide a loose upper bound.

\section{Technical Proofs} \label{sect:AppendixProof}

\allowdisplaybreaks

In this supplement, we provide detailed proofs of the theoretical analysis.
\begin{enumerate}[label=]
    \item \ref{sect:ConcenCoeff} Derivation of the Properties of the Concentratability Coefficient.
    \item \ref{sect:EstBound} Estimation Bound for the Low-Rank Optimization: Proof of Theorem \ref{thm:estimation_bound_informal}.
    \item \ref{sect:rtv_error} Estimation Bound for the Rotation-Truncation-Vectorization.
    \item \ref{sect:ConfBound} Confidence Bound of the reduced parameter: Proof of Lemma \ref{lem:confidence_bound}.
\end{enumerate}

\subsection{Properties of the Concentratability Coefficient} \label{sect:ConcenCoeff}
In this section, we derive the theoretical properties of the concentratability coefficient under rotation and truncation. Suppose $y_t = U^\top x_t$ for orthonormal matrix $U$. Let $\Sigma_x = \sum x_t x_t^\top$ and $\Sigma_y = \sum y_t y_t^\top$. Then,
$$
\Sigma_y = \sum U^\top x_t x_t^\top U = U^\top \Sigma_x U
$$
and therefore, $\Sigma_y = U^\top \Sigma_x^{-1} U$. Therefore, for $y = U^\top x$, $y^\top \Sigma_y^{-1} y = x^\top U U^\top \Sigma_x^{-1} U U^\top x = x^\top \Sigma_x^{-1} x$ and hence, the concentratability coefficient is constant over rotation.

The concentratability coefficient decreases as we remove covariates. Let $x^\top = (x_1^\top, x_2^\top)$. Also denote
$$
\Sigma = \sum x x^\top = \begin{pmatrix}
    \Sigma_{11} & \Sigma_{12} \\ \Sigma_{21} & \Sigma_{22}
\end{pmatrix}.
$$

Note that 
$$
\Sigma^{-1} = \begin{pmatrix}
    \Sigma_{11\cdot 2}^{-1} & -\Sigma_{11\cdot 2}^{-1} \Sigma_{12} \Sigma_{22}^{-1} \\
    -\Sigma_{22}^{-1} \Sigma_{21} \Sigma_{11\cdot 2}^{-1} & \Sigma_{22}^{-1} + \Sigma_{22}^{-1}\Sigma_{21} \Sigma_{11\cdot 2}^{-1} \Sigma_{12} \Sigma_{22}^{-1}
\end{pmatrix},
$$
where $\Sigma_{11\cdot 2} = \Sigma_{11} - \Sigma_{12}\Sigma_{22}^{-1}\Sigma_{21}$. Therefore we can verify that 
\begin{align*}
    x^\top \Sigma^{-1} x &= x_2^\top \Sigma_{22}^{-1} x_2 + (x_1 - \Sigma_{12} \Sigma_{22}^{-1} x_2)^\top \Sigma_{11\cdot 2}^{-1} (x_1 - \Sigma_{12} \Sigma_{22}^{-1} x_2) \\
    &\ge x_2^\top \Sigma_{22}^{-1} x_2.
\end{align*}

Moreover note that
\begin{align*}
    \vect(U^\top x (V^\top \phi)^\top) &= (V^\top \phi) \otimes (U^\top x) \\
    &= (V^\top \otimes U^\top) (\phi \otimes x) \\
    &= (V \otimes U)^\top (\phi \otimes x),
\end{align*}
where $V \otimes U$ will also be an orthonormal matrix. This indicates rotation on both sides in our propose RTV will not change the value of the concentratability coefficient.

\subsection{Proof of Estimation Bound} \label{sect:EstBound}

In this section, we provide the proof of Theorem \ref{thm:estimation_bound_formal}, which is the formal statement of Theorem \ref{thm:estimation_bound_informal}. First, we establish a one-step bound of Algorithm \ref{alg:fgd} (Lemma \ref{lem:onestep}). Then we verify the conditions required for the one-step bound such as the restricted strong convexity (RSC) condition and the bound on the gradient (Lemma \ref{lem:RSC}, \ref{lem:BG}). Finally we extend the one-step result to the converged estimator to complete the proof. First we state the formal estimation bound.

\begin{thm}[Formal Estimation Bound] \label{thm:estimation_bound_formal}
    Suppose Assumptions \ref{assm:bounded_features} and \ref{assm:cov} holds. Then we have the estimation bound:
    $$
    \|\hat\Theta - \Theta^*\|_F \le \frac{12 B_x B_\phi}{\gamma \lambda_{\min}(V)} \sqrt{\frac{(3\gamma \lambda_{\min}(V) + 32 + 16\gamma) \cdot \sigma_1 r\log(\frac{d_x + d_\phi}{\delta})}{2\sigma_r n}}
    $$
    with probability at least $1-\delta$, where $V = \frac{1}{n_0} \sum_{t=1}^{n_0} z_t z_t^\top =  \frac{1}{n_0} \sum_{t=1}^{n_0} (\phi_t \phi_t^\top) \otimes (x_t x_t^\top)$.
\end{thm}

We begin the proof by re-stating the results on convergence of factored gradient descent algorithm on general loss functions from \citet{zhang2023generalized}. Since the result on \citet{zhang2023generalized} is a general result for where the loss is a joint function of a low-rank matrix and a sparse tensor, we state the simplified result without the assumptions on the tensor portion. First we define the minimal Frobenius norm under rotation.

\begin{defn}
    For $Z_1, Z_2 \in \bbR^{d\times r}$, $d(Z_1,Z_2)$ is defined as the minimal Frobenius norm between $Z_1$ and $Z_2$ under rotation i.e.,
    $$
    d(Z_1, Z_2) = \min_{R \in \mathbb{Q}_r} \|Z_1 - Z_2R\|_F ,
    $$
    where $\mathbb{Q}_r$ is the space of $r$-dimensional rotation matrices, i.e., $\mathbb{Q}_r = \{ R \in \bbR^{r\times r}: R^\top R = R R^\top = I_r\}$.
\end{defn}

Next we use the following Lemma to illustrate the one-step rate decrease in terms of the new metric.

\begin{lem}[Lemma 1 from \citep{zhang2023generalized}]\label{lem:onestep}
    Suppose $\ell$ satisfies the restricted strong convexity and restricted smoothness i.e., for any matrices $\Theta_1, \Theta_2 \in \mathbb{B}(\Theta^*, 1)$ with rank at most $r$,
    $$
    \frac{\mu}{2} \|\Theta_2 - \Theta_1\|_F^2 \le \ell(\Theta_2) - \ell(\Theta_1) - \langle \nabla \ell(\Theta_1), \Theta_2 - \Theta_1 \rangle \le \frac{L}{2} \|\Theta_2 - \Theta_1\|_F^2 .
    $$
    Let $\sigma_1, \ldots, \sigma_r$ denote the singular values of $\Theta^*$. Let $c_1, c_2$ be a constant such that $c_1 \le \min\{1/32, \mu/(192L^2)\}$, $c_2 \le \sqrt{\min\{\mu,2\}/(6L+4)}$ and consider the step size $\delta = c_1/\sigma_1$. Let $Z^* = [U^* ; V^*]$, and $Z^{(t)} = [U^{(t)}, V^{(t)}$. If $d(Z^{(t)}, Z^*) \le c_2 \sqrt{\sigma_r^*}$, then Algorithm \ref{alg:fgd} satisfies
    $$
    d^2(Z^{(t+1)}, Z^*) \le \rho d^2(Z^{(t)}, Z^*) - \frac{\delta \mu}{4} \|\Theta^{(t)} - \Theta^*\|_F^2 + C_1 \|\nabla \ell(\Theta^*)\|_2^2 ,
    $$
    where $\rho = 1- \delta\mu \sigma_r/16$, $C_1 = 48r\delta^2 \sigma_1 + 2\delta(8r/\mu + r/L)$.
\end{lem}

Next step is checking the assumptions of Lemma \ref{lem:onestep}. Note that 
\begin{align*}
    |\theta^\top z_t| &\le \|\theta\|_2 \cdot \|z_t\|_2 \\
    &= \|\Theta\|_F \cdot \|Z_t\|_F \\
    &= \|\Theta\|_F \cdot \|\phi(s_t, a_{t1}) - \phi(s_t, a_{t2})\|_2 \cdot \|x_t\|_2 \\
    &\le 2B_{\theta} B_{\phi} B_x
\end{align*}

and therefore 
$$
\frac{1}{4} V \succeq \nabla_\theta^2 \calL(\theta) \succeq \gamma V,
$$

where $\gamma = (2+\exp(-2B_\theta B_\phi B_x) + \exp(2B_\theta B_\phi B_x))^{-1}$ and 
\begin{align*}
    V &= \frac{1}{n} \sum_{t=1}^n z_t z_t^\top  \\
    &= \frac{1}{n} \sum_{t=1}^n \left(\left( (\phi(s_t, a_{t1}) - \phi(s_t, a_{t0})) (\phi(s_t, a_{t1}) - \phi(s_t, a_{t0}) )^\top \right) \otimes (x_t x_t^\top)\right).
\end{align*}

Then we have the following result on the restricted strong convexity and smoothness of the loss function:

\begin{lem} [Restricted Strong Convexity and Smoothness] \label{lem:RSC}
$$
\gamma \lambda_{\min}(V) \|Y- X\|_F^2 \le {\calL}_n(Y) - {\calL}_n(X) - \langle \nabla {\calL}_n (X), Y-X \rangle \le \frac{1}{4} \lambda_{\max}(V) \|Y -X\|_F^2 ,
$$
where $\gamma := \frac{\exp(2B_\theta B_\phi B_x)}{(1 + \exp(2B_\theta B_\phi B_x))^2}$ is a constant and equivalently $X,Y \in \calB_F(\Theta^*, 1)$. 
\end{lem}

Next to utilize the result of Lemma \ref{lem:onestep}, we need a bound on the last term of the inequality, the gradient of the loss function evaluated in the true parameter. 

\begin{lem}[Bounded Gradient] \label{lem:BG}
    $$
    \left\| \frac{1}{n} \sum_{t=1}^n \left( \bbone(y_t=0) - \frac{\exp(-\theta^\top z_t)}{1 + \exp(-\theta^\top z_t)} \right) Z_t \right\|  \le 2B_x B_\phi \sqrt{\frac{\log(\frac{d_x+d_\phi}{\delta})}{n}} = \epsilon(n,\delta)
    $$
    with probability at least $1-\delta$.
\end{lem}

Plugging in the results of Lemma \ref{lem:RSC} and \ref{lem:BG} to Lemma \ref{lem:onestep} and since $\lambda_{\max}(V) \ge \lambda_{\min}(V)$., we have the following result on the converged estimator $\hat\Theta$:
\begin{cor}[Formal Estimation Bound] \label{cor:estimation_bound}
    If $\|\Theta_0 - \Theta^*\| \le c_2 \sqrt{\sigma_r}$ where $c_2 \le \sqrt{\min\{\mu, 2\} \cdot (6L + 4)}$, then:
    \begin{align*}
    \|\hat\Theta - \Theta^*\|_F^2 &\le \frac{288\sigma_1(3\gamma \lambda_{\min}(V) + 32 + 16\gamma) \cdot r \log(\frac{d_x+d_\phi}{\delta})}{4\sigma_r \gamma^2 \lambda_{\min}^2(V)} B_x^2 B_\phi^2
    \end{align*}
    or equivalently
    $$
    \|\hat\Theta - \Theta^*\|_F \le \frac{12B_x B_\phi}{\gamma \lambda_{\min}(V)} \sqrt{\frac{ (3\gamma \lambda_{\min}(V) + 32 + 16\gamma) \cdot \sigma_1 r \log(\frac{d_x+d_\phi}{\delta})} {2\sigma_r n} }.
    $$
\end{cor}

\subsubsection{Proof of Lemma \ref{lem:RSC}}

For proof of Lemma \ref{lem:RSC}, we use Taylor expansion and bounds on the second moment of the loss function. From the definition of the loss function, We can easily check that
$$
\nabla^2 {\calL} (\theta) = \frac{1}{n} \sum_{t=1}^n \frac{\exp(-\theta^\top z_t)}{(1+\exp(-\theta^\top z_t))^2} z_t z_t^\top .
$$

Further note that from the Taylor expansion, we have:
\begin{equation} \label{eqn:hessian}
{\calL}(Y) - {\calL}(X) - \langle \nabla {\calL} (X), Y-X \rangle = (y-x)^\top \nabla^2 {\calL}(\bar{\bar{\theta}}) (y-x),
\end{equation}
where $x = \vect(X), y = \vect(Y)$, $\bar{\theta} = cx + (1-c)y$ for some $c\in (0,1)$. Now assume $x, y, \in \mathcal{B}(\theta^*,1)$, and assume $\|\theta\|_2 \le B_\theta$ for all $\theta \in \mathcal{B}(\theta^*,1)$. Then, $\|\bar{\theta}\|_2 \le B_\theta$, so $|(\bar{\theta})^\top z_t| \le 2 B_\theta B_\phi B_x$. Then, we can easily check that 
$$
\gamma := \frac{\exp(2B_\theta B_\phi B_x)}{(1 + \exp(2B_\theta B_\phi B_x))^2} \le \frac{\exp(-(\bar\theta)^\top z_t)}{(1 + \exp(-(\bar\theta)^\top z_t))^2} \le \frac{1}{4}.
$$

Therefore,
$$
\gamma V \preceq \nabla_\theta^2 \mathcal{L}(\bar\theta) \preceq \frac{1}{4} V.
$$

Combined with (\ref{eqn:hessian}), we have
$$
\gamma \lambda_{\min}(V) \|Y-X\|_F^2 \le {\calL}(Y) - {\calL}(X) - \langle \nabla {\calL} (X), Y-X \rangle \le \frac{1}{4} \lambda_{\max}(V) \|Y-X\|_F^2,
$$

which completes the proof.

\subsubsection{Proof of Lemma \ref{lem:BG}}

We use the following Bernstein inequality for rectangular matrices from \citep{tropp2012user}:
\begin{lem}[(Bernstein Inequality for Rectangular Matrices, Theorem 1.6 of \citep{tropp2012user}] \label{lem:matrixbernstein}
    Suppose $M_k \in \bbR^{d_1\times d_2}$ are independent random matrices. Suppose $\bbE M_k = 0$ and $\|M_k\|_2 \le R$ almost surely. Define
    $$
    \sigma^2 := \max \left\{ \left\| \sum_k \bbE (M_k M_k^\top) \right\|, \left\|\sum_k \bbE(M_k^\top M_k) \right\| \right\} .
    $$
    Then for all $t\ge 0$,
    $$
    \bbP\left\{ \left\| \sum_k M_k \right\| \ge t \right\} \le (d_1 + d_2) \cdot \exp \left( \frac{-t^2/2}{\sigma^2 + Rt/3} \right) .
    $$
\end{lem}

For brevity, let us denote $p_t = \frac{1}{1+\exp(-\theta^\top z_t)}$ and $\epsilon_{t} =y_t - p_t$. We can easily check that $\bbE \epsilon_t = 0$. Denote $\phi_t = \phi(s_t, a_{t1}) - \phi(s_t, a_{t0})$ and plug in $M_t = \epsilon_t Z_t$ to apply Lemma $\ref{lem:matrixbernstein}$.

Now to verify $R$ in the conditions of Lemma \ref{lem:matrixbernstein}, note that $|\epsilon_{t}|\le 1$ and $\|Z_t\|_2 = \|x_t(\phi(s_t, a_{t1}) - \phi(s_t, a_{t2}))\|_2 \le 2 B_x B_\phi$. Therefore we can say $R = 2B_x B_\phi$.

Next we verify $\sigma^2$. Note that:
\begin{align*}
    \sum_{t=1}^n \bbE M_t M_t^\top &=  \bbE \left[ \sum_{t=1}^n \epsilon_{t}^2 \cdot x_t \phi_t^\top \phi_t x_t^\top \right] \\
    &= \sum_{t=1}^n \bbE \epsilon_{t}^2 \|\phi_t\|_2^2 \cdot x_t x_t^\top \\
    &= \sum_{t=1}^n p_t (1-p_t) \cdot \|\phi_t\|_2^2 \cdot x_t x_t^\top,
\end{align*}
since $\bbE \epsilon_{it}^2 = \Var(\epsilon_{it}) = \Var(y_{it}) = (\bbE y_{it})(1- \bbE y_{it})$. Also note that $\|xx^\top v\| = |x^\top v| \cdot \|x\| \le \|x\|^2 \cdot \|v\|$, so $\|xx^\top\|_2 = \|x\|^2$. Therefore,
\begin{align*}
    \left\| \sum_{t=1}^n \bbE \left[ M_t M_t^\top \right] \right\|_2  &= \left\| \sum_{t=1}^n p_t (1-p_t) \|\phi_t\|_2^2 \cdot x_t x_t^\top \right\|_2 \\
    &\le \frac{1}{4} \sum_{t=1}^n \|\phi_t\|_2^2 \cdot\|x_tx_t^\top\|_2 \\
    & = \frac{1}{4} \sum_{t=1}^n \|\phi_t\|_2^2 \cdot \|x_t\|_2^2 \\
    &\le n B_x^2 B_\phi^2 
\end{align*}

and similarly we have:
$$
\left\| \sum_{t=1}^n \bbE Z_t^\top Z_t \right\|_2 \le n B_x^2 B_\phi^2  .
$$

Now applying the Lemma \ref{lem:matrixbernstein} with $R= 2B_x B_\phi$ and $\sigma^2 = n B_x^2 B_\phi^2$ we have:
$$
\bbP \left( \left\|\frac{1}{n} \sum_{t=1}^n Z_t \right\|_2 \ge t \right) \le (d_x + d_\phi) \exp \left(-\frac{3nt^2}{B_x B_\phi(6B_xB_\phi + 4t)} \right) .
$$

Let $\eta = \log\frac{d_x+d_\phi}{\delta}$ for some $\delta \in (0,1)$. Assuming $n \ge \frac{16}{9}  \eta$, and select $t = 2B_x B_\phi \sqrt{\frac{\eta}{n}}$. Then we have $4t \le 6B_x B_\phi $. Therefore we have
\begin{align*}
\bbP \left( \left\|\frac{1}{n} \sum_{t=1}^n Z_t \right\|_2 \ge t \right) &\le (d_x + d_\phi) \exp \left(-\frac{3nt^2}{B_x B_\phi(6B_xB_\phi + 4t)} \right) \\ 
& \le (d_x + d_\phi) \exp \left( -\frac{ 12B_x^2 B_\phi^2 \eta }{12B_x^2 B_\phi^2 } \right) \\
&\le (d_x + d_\phi) \exp(-\eta) = \delta .
\end{align*}
Therefore,
$$
\left\| \frac{1}{n} \sum_{t=1}^n (y_t - p_t) \phi_t x_t^\top \right\|_2 \le 2B_x B_\phi \sqrt{\frac{\log(\frac{d_x+d_\phi}{\delta})}{n}} = \epsilon(n,\delta) ,
$$
with probability at least $1-\delta$. This completes the proof of Lemma \ref{lem:BG}.

\subsubsection{Proof of Corollary \ref{cor:estimation_bound}}

Before moving on to the main proof of Corollary \ref{cor:estimation_bound}, we first state the modified result from \citep{wang2017unified} and Lemma \ref{lem:onestep}.

\begin{prop}\label{prop:dist}
    For $i =1,2,$, let $U_i \in \bbR^{d_1\times r}$, $V_i \in \bbR^{d_2\times r}$ and $X_i = U_i V_i^\top$. Now denote
    $$
    Z_1 = \begin{pmatrix} U_1\\ V_1 \end{pmatrix}, \quad 
    Z_2 = \begin{pmatrix} U_2 \\ V_2 \end{pmatrix} .
    $$
    Then,
    $$
    \|X_1 - X_2\|_F^2 \le 2 (\|Z_2\|_2 + d(Z_1, Z_2))^2 d^2(Z_1, Z_2) .
    $$
\end{prop}

\begin{proof}
    Note that for two positive definite matrices $A,B$ with eigenvalue decomposition $A = P\Lambda P^\top$ and $B = QDQ^\top$, 
    \begin{align*}
        \tr(AB) &= \tr(P \Lambda P^\top QDQ^\top) \\
        &= \tr(Q^\top P \Lambda P^\top Q D) &(\because \tr(AB) = \tr(BA))\\
        &\le \tr(Q^\top P \Lambda P^\top Q) \cdot \|D\|_2 \\
        &= \tr(\Lambda) \cdot \|D\|_2 \\
        &= \tr(A) \cdot \|B\|_2 .
    \end{align*}

    This implies:
    \begin{align*}
        \|AB\|_F^2 &= \tr(ABB^\top A^\top) \\
        &= \tr(BB^\top A^\top A) \\
        &\le \tr(BB^\top) \|A^\top A\|_2 \\
        &= \|B\|_F^2 \|A\|_2^2
    \end{align*}
    
    and therefore $\|AB\|_F \le \|B\|_F \|A\|_2$. Now consider $R \in \mathbb{Q}_r$. Note that:
    \begin{align*}
    \|U_1(V_1- V_2R)^\top \|_F^2 &= \tr(U_1 (V_1 - V_2R)^\top (V_1 - V_2R) U_1^\top) \\
    &= \tr((V_1-V_2R)^\top (V_1 - V_2R) U_1^\top U_1) \\
    &\le \|U_1^\top U_1\|_2 \cdot \tr((V_1 - V_2R)^\top (V_1-V_2R)) \\
    &=  \|U_1\|_2^2 \|V_1 - V_2R\|_F^2 .
    \end{align*}
    Similarly we have $\|(U_1 - U_2R) R^\top V_2^\top\|_F^2 \le \|V_2\|_2^2 \|U_1 - U_2R\|_F^2$. Now note that for arbitrary vector $v$,
    $$
    \|Z_iv\|_2^2 = \|U_iv\|_2^2 + \|V_iv\|_2^2 \ge \|U_iv\|_2^2 .
    $$
    
    So we have $\|Z_i\|_2 \ge \|U_i\|_2$ and similarly $\|Z_i\|_2 \ge \|V_i\|_2$. Now note that:
    \begin{align*}
    \|Z_1\|_2 &\le \|Z_1 - Z_2R\|_2 + \|Z_2R\|_2 \\
    &= \|Z_1 - Z_2R \|_2 + \|Z_2\|_2 &(\because R \in \mathbb{Q}_r)\\
    &\le \|Z_1 - Z_2R\|_F + \|Z_2\|_2 . &(\because \|A\|_2 \le \|A\|_F)
    \end{align*}

    And this holds for arbitrary $R \in \mathbb{Q}_r$. Therefore, 
    \begin{equation} \label{eqn:rotineq}
    \|Z_1\|_2 \le \|Z_2\|_2 + d(Z_1, Z_2) .
    \end{equation}
    
    Therefore we have:
    \begin{align*}
        \|X_1 - X_2\|_F^2 &= \|U_1 V_1^\top - U_2 V_2^\top \|_F^2 \\
        &= \|U_1 V_1^\top - U_1 R^\top V_2^\top + U_1 R^\top V_2^\top - U_2 R R^\top V_2^\top \|_F^2 \\
        &\le (\|U_1(V_1 - V_2R)^\top \|_F+ \|(U_1 - U_2R) R^\top V_2^\top \|_F)^2 \\
        &\le 2(\|U_1(V_1 - V_2R)^\top \|_F^2 + \|(U_1 - U_2R)R^\top V_2^\top\|_F^2) &(\because (a+b)^2 \le 2(a^2+b^2))  \\
        &\le 2 (\|V_2\|_2^2\|U_1 - U_2R\|_F^2 + \|U_1\|_2^2\|V_1 - V_2R\|_F^2) &(\because \|AB\|_F^2 \le \|A\|_F\|B\|_2)\\
        &\le 2 (\|Z_2\|_2^2\|U_1 - U_2R\|_F^2 + \|Z_1\|_2^2\|V_1 - V_2R\|_F^2) \\
        &\le 2 (\|Z_2\|_2 + d(Z_1, Z_2))^2 (\|U_1 - U_2R\|_F^2 + \|V_1 - V_2R\|_F^2) &(\because (\ref{eqn:rotineq})\\
        &= 2 (\|Z_2\|_2 + d(Z_1, Z_2))^2  \|Z_1 - Z_2R\|_F^2 .
    \end{align*}
    Since this holds for arbitrary $R \in \mathbb{Q}_r$, The statement holds. 
\end{proof}

Now we can finally prove Corollary \ref{cor:estimation_bound}.

\begin{proof}
    Assume $d(Z^{(0)}, Z^*) \le \kappa_1$ and $C_1 \epsilon(n,\delta) \le (1-\rho) \kappa_1$. Then by induction and using Lemma \ref{lem:onestep}, we can check that $d(Z^{(t)}, Z^*) \le \kappa_1$ for every $t\ge 0$. Also we have:
    \begin{align*}
    d^2(Z^{(t)}, Z^*) &\le \rho^t d^2(Z^{(0)}, Z^*) + \frac{C_1(1-\rho^t)}{1-\rho} \epsilon^2(n,\delta) \\
    &\le \rho^t \kappa_1^2 + \frac{C_1}{1-\rho} \epsilon^2(n,\delta) .
    \end{align*}

    Note that for unit vector $v\in \bbR^{r}$ with $\|v\|_2 = 1$,
    \begin{align*}
        \|Z^*v\|_2^2 &= \left\| \begin{pmatrix} U^*v \\ V^*v \end{pmatrix} \right\|_2^2 \\
        &= \|U^*v\|_2^2 + \|V^*v\|_2^2  \\
        &\le \|U^*\|_2^2 + \|V^*\|_2^2 \\
        &= 2 \|\Theta^*\|_2.
    \end{align*}
    
    Therefore $\|Z^*\|_2 \le \sqrt{2\|\Theta^*\|_2} = \sqrt{2\sigma_1}$. Using Proposition \ref{prop:dist}, we have:
    \begin{align*}
        \|\Theta^{(t)} - \Theta^*\|_F^2 &\le 2(\sqrt{2\sigma_1} + d(Z^{(t)}, Z^*))^2 d^2(Z^{(t)}, Z^*)\\
        &\le (4\sigma_1 + 2 d^2(Z^{(t)}, Z^*)) d^2(Z^{(t)}, Z^*) . &(\because (a+b)^2 \le 2(a^2+b^2))
    \end{align*}
    
    As $t\to \infty$, we have:
    $$
    \|\hat{\Theta} - \Theta^*\|_F^2 \le \left( 4\sigma_1 + \frac{2C_1}{1-\rho} \epsilon^2(n,\delta) \right) \cdot \frac{C_1}{1-\rho} \epsilon^2(n,\delta) .
    $$

    Now assume $\epsilon^2(n,\delta) \le \frac{5(1-\rho)\sigma_1}{2C_1}$. Then we have:
    $$
    \|\hat\Theta - \Theta^*\|_F^2 \le \frac{9\sigma_1 C_1}{1-\rho} \epsilon^2(n,\delta) .
    $$

    Now note that $C_1 = r(48\delta^2\sigma_1 + 2\delta(8/\mu + 1/L))$. So we have:
    $$
    \|\hat\Theta - \Theta^*\|_F^2 \le \frac{9\sigma_1}{1-\rho}\left(48\delta^2 \sigma_1 + 2\delta\left( \frac{8}{\mu} + \frac{1}{L}\right) \right) r \epsilon^2(n,\delta) .
    $$

    Plugging in the values of $\delta = c_1/\sigma_1$ and $\rho = 1 - \delta\mu\sigma_r/16$ we have:
    $$
    \|\hat\Theta - \Theta^*\|_F^2 \le 9\sigma_1 \cdot \frac{32}{\mu\sigma_r} \left( 24c_1 + \frac{8}{\mu} + \frac{1}{L} \right) r \epsilon^2(n,\delta) .
    $$

    Now note that the constants are defined as:
    \begin{align*}
        \mu &= \gamma \lambda_{\min}(V) \\
        L &= \frac{1}{4} \lambda_{\max}(V) \\
        \epsilon(n,\delta) &= \sqrt{\frac{\log(\frac{d_x+d_\phi}{\delta})}{n}} B_x B_\phi \\
        c_1 &\le \min \left( \frac{1}{32}, \frac{\mu}{192L^2} \right) \le \frac{1}{32} .
    \end{align*}

    Plugging in the values we achieve:
    \begin{align*}
    \|\hat\Theta - \Theta^*\|_F^2 &\le \frac{288\sigma_1}{\sigma_r} \cdot \frac{n}{\gamma \lambda_{\min}(V)} \cdot \left( 24 c_1 + \frac{8}{\gamma \lambda_{\min}(V)} + \frac{4}{\lambda_{\max}(V)} \right) r \cdot \frac{\log(\frac{d_x+d_\phi}{\delta})}{n} B_x^2 B_\phi^2 \\
    &= \frac{288\sigma_1}{\sigma_r\cdot \gamma} \cdot \left( \frac{3}{4} + \frac{8}{\gamma \lambda_{\min}(V)} + \frac{4}{\lambda_{\max}(V)} \right) \frac{r\log(\frac{d_x+d_\phi}{\delta})}{\lambda_{\min}(V)} B_x^2 B_\phi^2.
    \end{align*}
    Plugging in $\lambda_{\max}(V) \ge \lambda_{\min}(V)$ completes the proof.

\end{proof}

\subsection{Estimation Error from Rotation-Truncation-Vectorization} \label{sect:rtv_error}

Here we show the estimation bound of ``rotation-truncation'' with respect to the estimated subspace. Consider the following Lemma from \citet{jun2019bilinear}.

\begin{lem}[Wedin's sin$\Theta$ theorem, Appendix B of \citep{jun2019bilinear}] \label{lem:wedin}
    Let $\hat{\Theta} = [\hat{U}, \hat{U}_\perp] \hat{D} [\hat{V}, \hat{V}_\perp]^\top$ and $\Theta^* = U^* D^* V^*$ be the SVD of $\hat\Theta$ and $\Theta^*$. Then,
    $$
    \|\hat{U}_\perp^\top U^* \|_F \|\hat{V}_\perp^\top V^*\|_F \le \frac{\|\hat{\Theta} - \Theta^*\|_F^2}{\sigma_r^{*2}} ,
    $$
    where $\sigma_k^*$ is the $k$-th(smallest) eigenvalue of $\Theta^*$. 
\end{lem}

Now let $\Theta^*_r = \hat{U} \Theta^* \hat{V}^\top$ be the rotated true parameter with respect to the estimated subspace $\hat{U}, \hat{V}$. Let $(\Theta^*_r)_{22}$ denote the lower right $(d_x-r) \times (d_\phi-r)$ sub-matrix of $\Theta_r^*$. Then we have
$$
\|(\Theta^*_r)_{22}\|_F \le \|\hat{U}_\perp U\|_F \|\hat{V}_\perp V\|_F \cdot \|D^*\|_2 \le \frac{\sigma_1^*}{\sigma_r^{*2}} \|\hat{\Theta} - \Theta^*\|_F^2 .
$$

Combining this with our estimation bound from Theorem \ref{thm:estimation_bound_informal}, we have
\begin{equation}
\|(\Theta^*_r)_{22}\|_F \le  C_2 \cdot  \frac{\sigma_1^2 r\log(\frac{d_x + d_\phi}{\delta})}{\sigma_r^3 n_0} =: S_{\perp}.
\end{equation}

With this result, we can derive the estimation bound of $\hat{\theta}_{rtv}$ with respect to the semi-norm $\|\cdot\|_W$, which is used to define the confidence set of Algorithm \ref{alg:prs}.

\subsection{Proof of Lemma \ref{lem:confidence_bound}}\label{sect:ConfBound}

Consider the function 
$$
J(\theta_{rtv}) = \sum_{t=1}^n \left( \frac{1}{1+\exp(-z_{t,rtv}^\top \theta_{rtv})} - \frac{1}{1+ \exp(-\langle Z_t, \Theta^*\rangle)} \right) z_{t,rtv}.
$$

Further consider the likelihood function on the reduced space:
$$
{\calL}_{rtv}(\theta_{rtv}) = \frac{1}{n} \sum_{t=1}^n \left( \log \left( 1 + \exp(-z_{t,rtv}^\top \theta_{rtv})) \right) + (1-y_t) z_{t,rtv}^\top \theta_{rtv}) \right)
$$

Note that the gradient will be:
$$
\nabla {\calL}_{rtv}(\theta_{rtv}) = \frac{1}{n} \sum_{t=1}^n \left( \frac{1}{1 + \exp(-z_{t,rtv}^\top \theta_{rtv})} - y_t \right) z_{t,rtv}.
$$

So for the minimizer $\hat{\theta}_{rtv}$ of $\calL_{rtv}$, we have $\nabla \calL_{rtv}(\hat\theta_{rtv}) = 0$. Therefore, $J(\hat{\theta}_{rtv}) = \sum_{t=1}^n \epsilon_t z_{t,rtv}$ where $\epsilon_t = y_t - \bbE y_t$. Let $V\in \bbR^n$ be a vector with $\epsilon_t$ as its components, and let $X\in \bbR^{n\times k}$ be a matrix with $z_{t,rtv}$ as its rows. Then, $J(\hat\theta_{rtv}) = X^\top V$.

Therefore, 
\begin{align*}
    \|J(\hat\theta_{rtv})\|_{W^{-1}} = V^\top (X W^{-1} X^\top) V =: V^\top M V
\end{align*}
and we can easily check that $\tr(M)\le nk, \tr(M^2) \le n^2 k$ and $\|M\|\le n$. Using Berstein's inequality for sub-Gaussian random variables in quadratic form \citep{hsu2012tail}, we have
$$
 \|J(\hat\theta_{rtv})\|_{W^{-1}}^2 = V^\top M V \le C_1 \cdot n \cdot (k + \log(1/\delta)).
$$

Now note that 
$$
J(\theta^*_{rtv}) = \sum_{t=1}^n (\Phi(z_{t,rtv}^\top \theta^*_{rtv}) - \Phi(\langle Z_t, \Theta^*\rangle)) z_{t,rtv},
$$

where $\Phi(x) = 1/(1+e^{-x})$. Also note that $|\Phi'(x)| = e^{-x}/(1+e^{-x})^2 \le \frac{1}{4}$. Therefore,
\begin{align*}
    \Phi(z_{t,rtv}^\top \theta^*_{rtv}) - \Phi(\langle Z_t, \Theta^*\rangle)
    & \le \frac{1}{4}  |z_{t,rtv}^\top \theta^*_{rtv} - \langle Z_t, \Theta^*\rangle| \\
    &\le \frac{1}{4}  | \langle (Z_{t,r})_{22}, (\Theta^*_r)_{22} \rangle | \\
    &\le \frac{1}{4}  \|Z_t\|_F \cdot \|(\Theta^*_r)_{22}\|_F \\
    &\le \frac{1}{2} B_x B_\phi \cdot S_\perp
\end{align*}
and 
\begin{align*}
    \|J(\theta^*_{rtv})\|_{W^{-1}} &\le \frac{1}{2} B_x B_\phi \cdot S_\perp \cdot \sqrt{\left(\sum_{t=1}^n |z_{t,rtv}| \right)^\top W^{-1}  \left(\sum_{t=1}^n |z_{t,rtv}| \right)}\\
    &\le  \frac{1}{2} B_x B_\phi \cdot S_\perp \cdot n\sqrt{k}
\end{align*}
as
\begin{align*}
    \left(\sum_i z_i \right)^\top \cdot \left( \frac{1}{n} \sum_i z_i z_i^\top \right)^{-1} \left( \sum_i z_i \right) &= \tr \left[ \left( \frac{1}{n} \sum_i z_i z_i^\top \right)^{-1} \left( \sum_i z_i \right) \left(\sum_i z_i \right)^\top \right] \\
    &\le \tr \left[ \left(\frac{1}{n} \sum_i z_i z_i^\top \right)^{-1} \cdot n \cdot \left( \sum_i z_i z_i^\top \right) \right] \\
    &= n^2 k.
\end{align*}

Also note that
\begin{align*}
    &{\calL}_{rtv}(\hat{\theta}_{rtv}) - {\calL}_{rtv}(\theta^*_{rtv}) - \langle \nabla {\calL}_{rtv}(\theta^*_{rtv}) , \hat{\theta}_{rtv} - \theta^*_{rtv} \rangle \\
    =& (\hat{\theta}_{rtv} - \theta^*_{rtv})^\top \nabla^2 {\calL}_{rtv}(\bar{\theta}_{rtv}) (\hat{\theta}_{rtv} - \theta^*_{rtv}) \\
    \ge& (\hat{\theta}_{rtv} - \theta^*_{rtv})^\top \cdot \frac{\gamma}{n}\cdot \sum_{t=1}^n z_{t,rtv} z_{t,rtv}^\top (\hat{\theta}_{rtv} - \theta^*_{rtv}) \\
    =&\gamma \cdot \|\hat{\theta}_{rtv} - \theta^*_{rtv}\|_W^2,
\end{align*}
where $W = X^\top  X$. Also note that $\calL_{rtv}(\hat\theta_{rtv}) \le \calL_{rtv}(\theta^*_{rtv})$. Therefore,

$$
\gamma \cdot \|\hat{\theta}_{rtv} - \theta^*_{rtv}\|_W^2 \le | \langle \nabla {\calL}_{rtv}(\theta^*_{rtv}) , \hat{\theta}_{rtv} - \theta^*_{rtv} \rangle | \le \|\nabla {\calL}_{rtv}(\theta^*_{rtv})\|_{W^{-1}} \cdot \|\hat{\theta}_{rtv} - \theta^*_{rtv}\|_{W}
$$
and 
\begin{align*}
\|\hat{\theta}_{rtv} - \theta^*_{rtv}\|_W &\le \frac{1}{\gamma} \|\nabla {\calL}_{rtv}(\theta^*_{rtv})\|_{W^{-1}} \\
&= \frac{1}{n\gamma} \|J(\hat\theta_{rtv}) - J(\theta^*_{rtv})\|_{W^{-1}} \\
&\le \frac{1}{n\gamma} \left( \|J(\hat\theta_{rtv})\|_{W^{-1}} + \|J(\theta^*_{rtv})\|_{W^{-1}} \right) \\
&\le \frac{1}{\gamma} \left( \sqrt{C_1 \cdot \frac{k + \log(1/\delta)}{n}} +  \frac{1}{2} B_x B_\phi \cdot S_\perp \cdot \sqrt{k} \right) 
\end{align*}

which completes the proof.

\subsection{Proof of Theorem \ref{thm:subopt}}

The key idea of the proof of Theorem \ref{thm:subopt} is decomposing the sub-optimality gap into three parts. For brevity, let us denote $r_{\theta}(x,s,\pi(x,s))$ as $r_{\theta}(x,s,\pi)$. Now define the pessimistic value function $\hat{J}$ as:
$$
\hat{J}(\pi) := \min_{\|\theta_{rtv} - \hat\theta_{rtv}\|_W\le C} {\bbE}_{(x,s) \sim \rho} r_{\theta_{rtv}}(x,s,\pi).
$$

Then the proposed pessimistic policy $\hat\pi$ is given as: $\hat\pi := \argmax_{\pi}\hat{J}(\pi)$. Let us first consider the individualized policy where each policy is given in the form of $\pi(x,s)$. Note that the sub optimality gap can be decomposed as sum of three differences:
$$
J(\pi^*) - J((\hat\pi) = J(\pi^*) - \hat{J}(\pi^*) + \hat{J}(\pi^*) - \hat{J}(\hat\pi) + \hat{J}(\hat\pi) - J(\hat\pi).
$$

By the definition of $\hat\pi$, the second difference satisfies $\hat{J}(\pi^*) - \hat{J}(\hat\pi) \le 0$. Further note that for arbitrary $x,s,a$,
\begin{align*}
    \left| r_{\Theta^*}(x,s,a) - r_{\theta^*_{rtv}}(x,s,a) \right| \le S_\perp \cdot \|z_{rtv}\| \le B_x B_\phi S_\perp.
\end{align*}

Then conditioned on $\|\theta^*_{rtv} - \hat\theta_{rtv}\|_W \le C$, the third difference satisfies:
\begin{align*}
    \hat{J}(\hat\pi) - J(\hat\pi) &= \min_{\|\theta_{rtv} - \hat\theta_{rtv}\|_W\le C} {\bbE}_{(x,s)\sim \rho} r_{\theta_{rtv}}(x,s, \hat\pi) - {\bbE}_{(x,s)\sim\rho} r_{\Theta^*}(x,s,\hat\pi) \\
    &\le B_x B_\phi S_\perp +  \min_{\|\theta_{rtv} - \hat\theta_{rtv}\|_W\le C} {\bbE}_{(x,s)\sim \rho} r_{\theta_{rtv}}(x,s, \hat\pi) - {\bbE}_{(x,s)\sim\rho} r_{\theta^*_{rtv}}(x,s,\hat\pi) \\
    &\le B_x B_\phi S_\perp.
\end{align*}

As for the first difference:
\begin{align*}
    J(\pi^*) - \hat{J}(\pi^*) &= {\bbE}_{(x,s)\sim\rho} r_{\Theta^*}(x,s,\pi^*) - \min_{\|\theta_{rtv} - \hat\theta_{rtv}\|_W\le C} {\bbE}_{(x,s)\sim \rho} r_{\theta_{rtv}}(x,s,\pi^*) \\
    &\le B_x B_\phi S_\perp + \max_{\|\theta_{rtv} - \hat\theta_{rtv}\|_W\le C} {\bbE}_{(x,s)\sim\rho} \left( r_{\theta^*_{rtv}}(x,s,\pi^*) - r_{\theta_{rtv}}(x,s,\pi^*) \right) \\
    &= B_x B_\phi S_\perp + \max_{\|\theta_{rtv} - \hat\theta_{rtv}\|_W\le C} \left( (\theta^*_{rtv} - \theta_{rtv})^\top {\bbE}_{(x,s)\sim\rho} z_{rtv}(x,s,\pi^*) \right) \\
    &\le B_x B_\phi S_\perp + \max_{\|\theta_{rtv} - \hat\theta_{rtv}\|_W\le C} \|\theta^*_{rtv} - \hat\theta_{rtv}\|_W \cdot \left\|  {\bbE}_{(x,s)\sim\rho} z_{rtv}(x,s,\pi^*) \right\|_{W^{-1}} \\
    &\le B_x B_\phi S_\perp + 2 C \cdot \left\|  {\bbE}_{(x,s)\sim\rho} z_{rtv}(x,s,\pi^*) \right\|_{W^{-1}}.
\end{align*}

Combining these results, we have the bound
$$
J(\pi^*) - J(\hat\pi) \le 2 B_x B_\phi S_\perp + 2C \cdot \left\|  {\bbE}_{(x,s)\sim\rho} z_{rtv}(x,s,\pi^*) \right\|_{W^{-1}}
$$
with probability at least $1-\delta$. Denote $C^* = \left\|  {\bbE}_{(x,s)\sim\rho} z_{rtv}(x,s,\pi^*) \right\|_{W^{-1}}$. Plugging in the confidence bound on $C$ from Lemma \ref{lem:confidence_bound}, we have

$$
J(\pi^*) - J(\hat\pi) \le 2B_x B_\phi S_\perp + \frac{2C^*}{\gamma} \left( \sqrt{C_3 \cdot \frac{k + \log(1/\delta)}{N_1}} +  \frac{1}{2} B_x B_\phi \cdot S_\perp \cdot \sqrt{k} \right).
$$

Note that we have a bound on $S_\perp$  given as:
$$
S_\perp =  C_2 \cdot  \frac{\sigma_1^2 r\log(\frac{d_x + d_\phi}{\delta})}{\sigma_r^3 N_0}
$$
from (\ref{eqn:S_perp}). Plugging in $n_0 = \Omega(\sqrt{n})$ gives us the optimality gap rate 
$$
J(\pi^*) - J(\hat\pi) = O\left( C^* \cdot \sqrt{\frac{(d_x+d_\phi)\cdot r + \log(1/\delta)}{N-N_0}}\right).
$$

Now assuming $N_1 = N-N_0 = \Omega(N)$ (For example, $N_0 = N/2$), we have the optimal rate in Theorem \ref{thm:subopt}.

\end{document}